\documentclass[12pt]{iopart}

\usepackage{harvard}
\usepackage{graphicx}
\usepackage{amsfonts}
\usepackage{hyperref}
\usepackage{float}
\usepackage{subfigure}
\begin{document}

\title[Functional data learning using convolutional neural networks]{Functional data learning using convolutional neural networks}

\author{J Galarza, T Oraby}

\address{School of Mathematical and Statistical Sciences, The University of Texas Rio Grande Valley, Edinburg, TX 78539, USA}
\vspace{10pt}
\begin{indented}
\item[]September 2023
\end{indented}

\begin{abstract}
In this paper, we show how convolutional neural networks (CNN) can be used in regression and classification learning problems of noisy and non-noisy functional data. The main idea is to transform the functional data into a 28 by 28 image. We use a specific but typical architecture of a convolutional neural network to perform all the regression exercises of parameter estimation and functional form classification. First, we use some functional case studies of functional data with and without random noise to showcase the strength of the new method. In particular, we use it to estimate exponential growth and decay rates, the bandwidths of sine and cosine functions, and the magnitudes and widths of curve peaks. We also use it to classify the monotonicity and curvatures of functional data, algebraic versus exponential growth, and the number of peaks of functional data. Second, we apply the same convolutional neural networks to Lyapunov exponent estimation in noisy and non-noisy chaotic data, in estimating rates of disease transmission from epidemic curves, and in detecting the similarity of drug dissolution profiles. Finally, we apply the method to real-life data to detect Parkinson's disease patients in a classification problem. The method, although simple, shows high accuracy and is promising for future use in engineering and medical applications.
\end{abstract}

\vspace{2pc}

\section{Introduction}

Functional data (FD) are functions observed for each unit over certain intervals; see \cite{xiaoying2021research,Ramsay2005Int}. FD appears in various scientific fields, such as engineering, geology, biology, medicine, pharmacology, and chemistry. It involves analyzing data in the form of continuous vector functions or curves, which could be treated as realizations of stochastic processes; see \cite{xiaoying2021research,yarger2022functional}. Functional data analysis (FDA) provides methods for extracting intrinsic information from infinite-dimensional and irregular observation data; see \cite{xiaoying2021research,Ramsay2005Int}. FDA combines statistics, spatial analysis, and multivariate modeling tools to analyze and predict functional data. It provides advantages over traditional pointwise estimation methods by using irregularly sampled data in space, time, and depth to fit space-time functional models \cite{gorecki2018selected}. 

Earlier stages of FDA were developed by \cite{ramsay1991kernel} to study the relationship between the ability of an examinee and his or her probability of correctly selecting an option in a standardized item response test. First, \cite{ramsay1991kernel} used kernels to correctly fit the observations to one dimension and later extended the idea in \cite{ramsay1995similarity} to fit the data to larger dimensions. Then, \cite{ramsay1996principal} introduced Principal Differential Analysis to find an approximate solution that solves the differential equation $Lu=0$ in which we can pick the order of the differential operator $L$ and a basis for the weights, using splines or other adequate functions. This technique was applied to approximate Chinese writing in \cite{ramsay2000functional} using a second-order operator. Later, complexity was added to this type of multilevel model \cite{ramsay2002multilevel}. For more information and applications of FD, we recommend reading the full work on Functional Analysis case studies of Ramsay's work in \cite{ramsay2002applied} and finding more mathematical foundation in \cite{Ramsay2005}.

 Functional data learning (FDL) has also received a lot of attention recently in the domain of machine learning and deep learning. Different approaches were used to establish predictive models of functional data like in \cite{zhao2012functional} where gradient descent was derived for classical Neural Networks using Fréchet derivates and the Reisz representation theorem to establish a deep learning method for FDL. In \cite{abraham2014machine}, the support vector machine method was applied to feature vectors obtained by voxels transformation. In \cite{pfisterer2019benchmarking}, some FDA R libraries were compared to machine learning techniques such as random forest and were shown to be outperformed by the latter. In \cite{zhang2021dynamic}, a new methodology was used to learn the bases of different functional subspaces to model functional data before applying learning methods. In \cite{basna2021machine}, orthogonal B-spline bases for FDA were shown to be more efficient than other Fourier-based methods.  
 
 Other deep learning and machine learning approaches include the work in \cite{yao2021deep}, in which inputs are fed directly into a layer composed of nodes of nested neural networks. The output of these neural networks is the basis function themselves. Inspired by the square root velocity, the characteristics in \cite{rafajlowicz2021learning} were obtained from functional data using derivatives. It was found to work well along with classification methods such as logistic regression and support vector machine. Other methods of FDL such as manifold learning have recently been introduced, see, e.g. \cite{hernandez2020high,mughal2020learning,hernandez2021functional}.

Convolutional neural network (CNN) has been used extensively in image recognition. The architecture of Neural Networks has been evolving since the introduction of very deep CNNs \cite{simonyan2014very}, ResNets \cite{he2016deep}, and MobileNets \cite{howard2017mobilenets,sandler2018mobilenetv2}. The efficiency of CNN is affected by its architecture and hardware \cite{polson2020deep}, as well as training and cross-validation sets and scaling \cite{tan2019efficientnet}. Several other advances have been made by adding other algorithms to CNN, such as the "You Only Look Once" approach proposed by \cite{redmon2016you,redmon2017yolo9000}. To our knowledge, convolutional neural networks have not been used before for functional data learning.

In this paper, we explore how to use CNNs in functional data learning. In Section 2, we introduce our method and the CNN architecture. In most of the problems, we use the same CNN architecture except for a few that will be described in place of their application. In Section 3, we start with some regression and classification problems of curves that represent different functions with and without random noise. In particular, we examine the ability of CNN to estimate parameters of exponential and trigonometric functions. In addition, we employed the same CNN to estimate the magnitude and width of the curve peaks. We also use CNN for the classification problems of increasing versus decreasing function, concave versus convex function, algebraic versus exponential growth, and discerning curves of 0, 1, and 2 peaks. In Section 4, we examine the ability of CNN to estimate parameters of characteristics of dynamical systems. In particular, we use it to estimate Lyapunov exponents of some chaotic curves with and without random noise. We apply the same CNN to estimate transmission rates from epidemic curves. Then, we applied a Siamese CNN to test the similarity between two drug dissolution curves (profiles). In Section 5, we apply the same methods to classify the actual functional data of cases of Parkinson's disease and control.

\section{Methods}
\begin{figure}[htp]
    \centering
    \includegraphics[width=15cm]{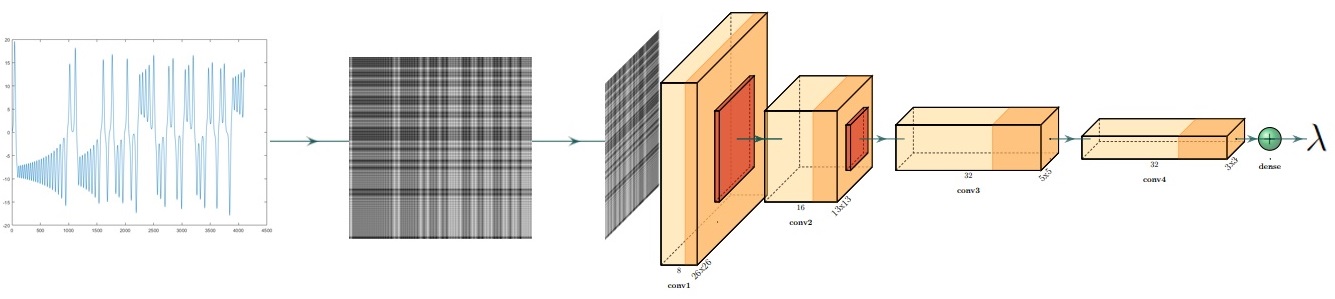}
    \caption{Diagram and procedure proposed for the regression and classification problems.}
    \label{fig:figure1}
\end{figure}

Let $\{(x_i,f(x_i)):i=1,2,\ldots, n\}$ be the data points of a graph with equidistance sampled points $x_i$, $i=1,2,\ldots, n$. The data pre-processing step in functional data learning via CNNs creates an input image using functional data points. We assume that $f$ is Min-Max normalized, so we can assume that $f:\mathbb{R}\to [0,1]$. A signed distance matrix $D$ is defined, so its elements $(i,j)$, $d_{ij}=f(x_i)-f(x_j)$. The matrix $D$ represents a grayscale intensity value that is used to produce a 28 by 28 grayscale image; see Figure \ref{fig:figure1}. 

Next, we describe a typical architecture as presented in Mathworks' documents \cite{MATLAB:2022}, which is based on Lenet networks in \cite{lecun1998gradient}. The first multilayer of CNN consists of batch normalization, a RELU activation function, and an average normalization layer. The first output is $13\times 13 \times 8 $ layers. The second multilayer is the same as the first with an output of $5\times 5 \times 16$ layers. The third multilayer does not contain an average pooling layer and has an output of $ 3\times 3 \times 32 $ layers. The final multilayer has no average pooling layer; however, it has a dropout layer of 20\% with a fully connected layer followed by a regression/classification layer. All codes and simulations were performed in MATLAB and using its deep learning toolbox \cite{MATLAB:2022}.

We use the same CNN's architecture for all of the regression and classification problems. We tested the procedure on different functional types and we start first by discussing our findings for seven case studies of functional data. 

\section{Results}
In this section, we discuss various regression and classification problems of functional data with and without random noise. We use randomly generated values for the functional parameters from specified ranges to produce training, validation, and testing data sets of sizes 1000, 100, and 100, respectively. We use $n=100$ data points for each curve. In the case of width estimation, height estimation, and number of peaks classification, we use $10000$ curves for training, and we use $n=1000$ functional data points for each curve. This would ensure that there will be a wide variety of heights and widths and enough peaks for each classification.

\subsection{Regression Problems}

\subsubsection{Exponential Function}\label{exp}
\begin{figure}[H]
    \centering
    \subfigure[]{\includegraphics[width=5.6cm]{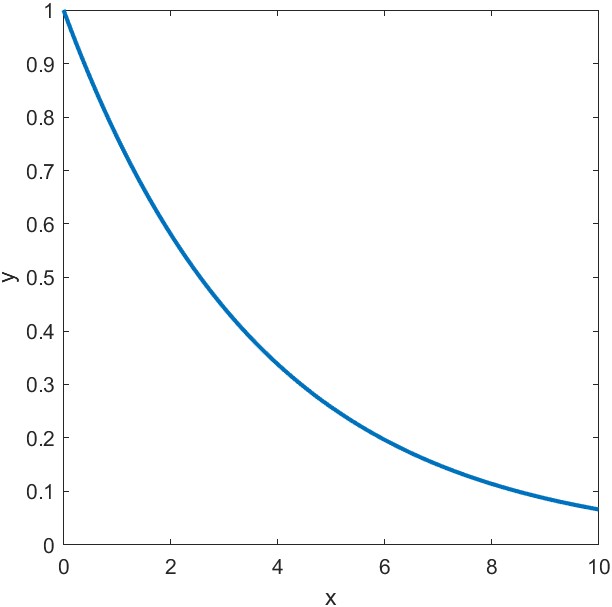}}\subfigure[]{\includegraphics[width=5.5cm]{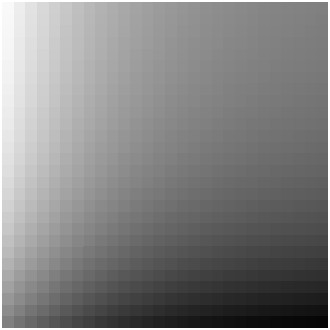}}
    \subfigure[]{\includegraphics[width=5.6cm]{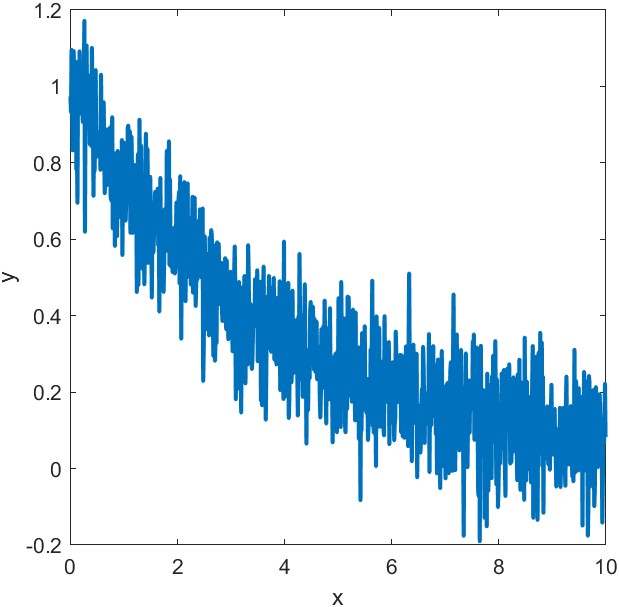}}\subfigure[]{\includegraphics[width=5.5cm]{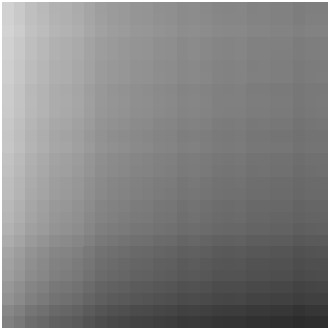}}    \subfigure[]{\includegraphics[width=5.5cm]{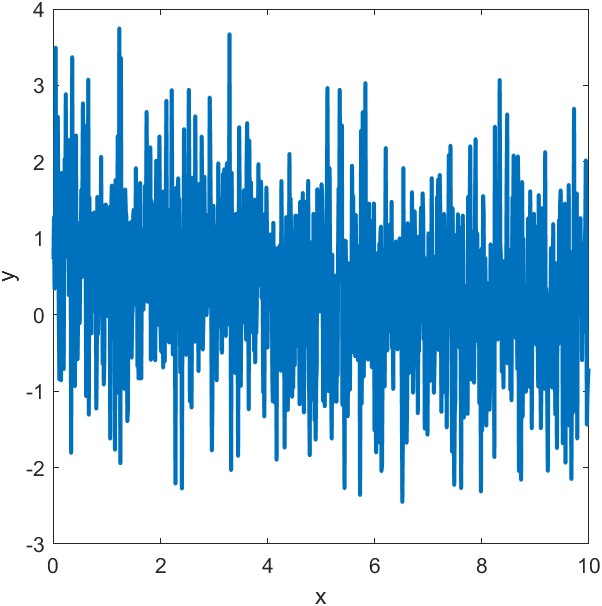}}\subfigure[]{\includegraphics[width=5.5cm]{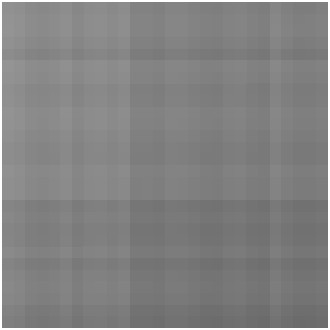}}
    \caption{(a) The curve of $y=\exp(\omega x)$ when $\omega=-.27$. (b) The 28 by 28 image that corresponds to the curve in (a). (c) The curve of $y=\exp(\omega x)+\sigma z$ when $\omega=-.27$ and $\sigma=.1$ where $z$ is a standard normal random variable. (d) The 28 by 28 image that corresponds to the curve in (c). (f) The curve of $y=\exp(\omega x)+\sigma z$ when $\omega=-.27$ and $\sigma=1$ where $z$ is a standard normal random variable. (f) The 28 by 28 image that corresponds to the curve in (e).}
    \label{fig:figure2}
\end{figure}

Let the exponential curve be $y=\exp(\omega x)$ with parameter $\omega$ representing the rate at which $y$ increases or decreases. The first task is to use our method to estimate $\omega$. See Figure \ref{fig:figure2} for the curve of $y=\exp (-0.27x)$  with and without noise as an example and the corresponding 28 by 28 image. For training, validation, and testing data, the parameter $\omega$ is sampled from a uniform distribution over the interval $[-1,1]$ for $1000$, $100$, and $100$ times, respectively.

\begin{figure}[H]
    \centering
    \subfigure[]{\includegraphics[width=7cm]{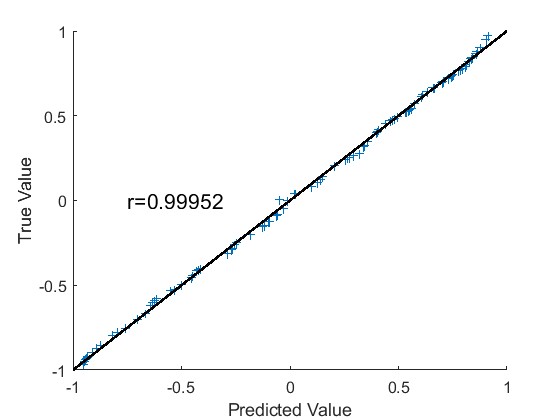}}  
    \subfigure[]{\includegraphics[width=7cm]{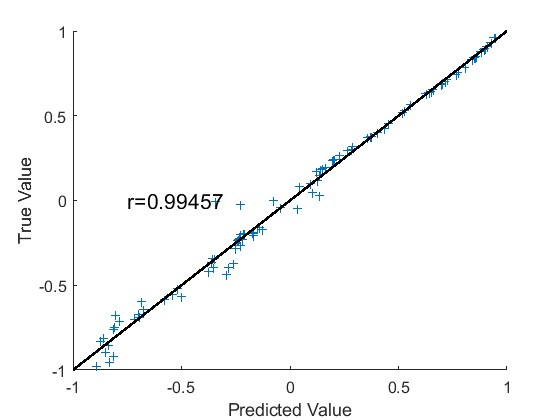}}
    \subfigure[]{\includegraphics[width=7cm]{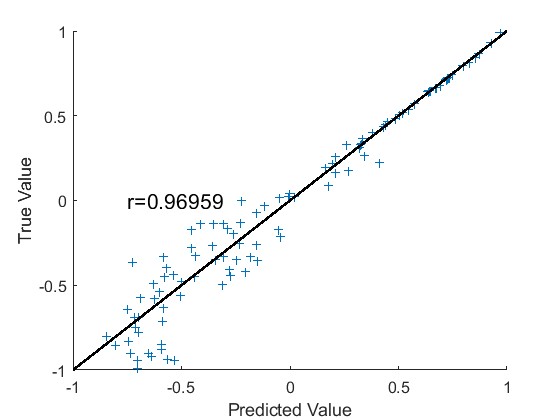}}
    \caption{Results of the exponential function regression using the test dataset without noise (a), with noise of magnitude $\sigma=.1$ (b), and with noise of magnitude $\sigma=1$ (c).}
    \label{fig:figure3}
\end{figure}

Figure \ref{fig:figure3} shows the predicted values versus the estimated value of the exponential function parameter that closely follows the diagonal line with no intercept and slope of one. Table \ref{tab:exp-results} shows a strong diagonal linear relationship between the true and predicted values of the rate.

\begin{table}[ht]
\centering
\caption{Correlation coefficient, intercept, and slope of a simple regression line of the actual and predicted values, and their p-values for three cases.}
\begin{tabular}{|c|c|c|c|}
\hline
Case & Correlation Coefficient ($r$) & Intercept (p-value) & Slope (p-value) \\
\hline
Without Noise & 0.999 & -0.000 (0.867) & 0.994 (0.049) \\
\hline
With Noise ($\sigma=0.1$) & 0.995 & 0.002 (0.692) & 0.995 (0.657) \\
\hline
With Noise ($\sigma=1$) & 0.970 & -0.011 (0.415) & 1.012 (0.655) \\
\hline
\end{tabular}
\label{tab:exp-results}
\end{table}

\subsubsection{Sine and Cosine Functions}
\begin{figure}[H]
    \centering
    \subfigure[]{\includegraphics[width=5.6cm]{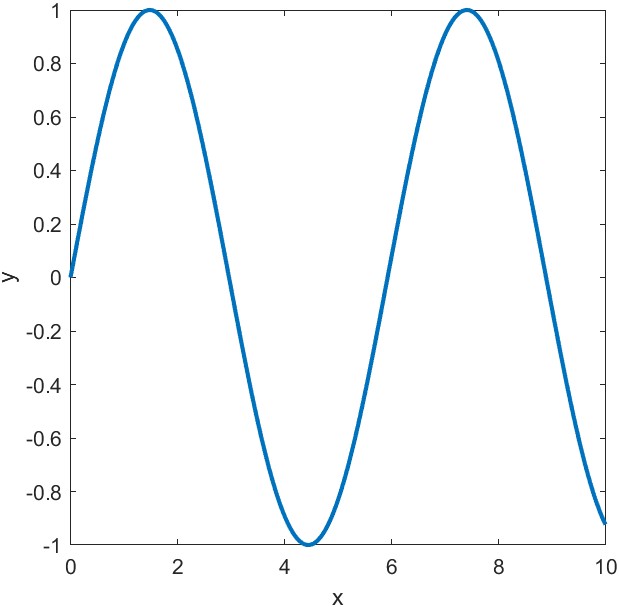}}
    \subfigure[]{\includegraphics[width=5.5cm]{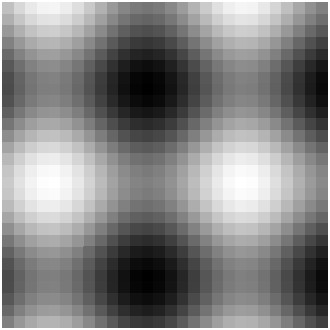}}
    \subfigure[]{\includegraphics[width=5.6cm]{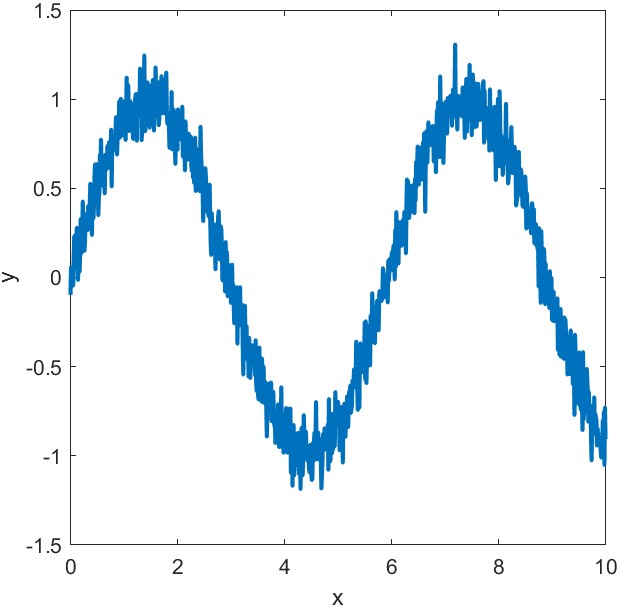}}\subfigure[]{\includegraphics[width=5.5cm]{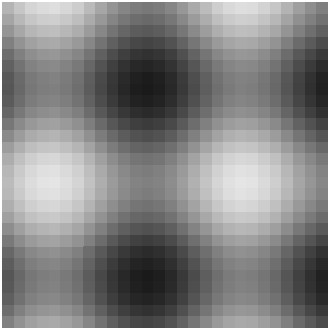}}
    \subfigure[]{\includegraphics[width=5.6cm]{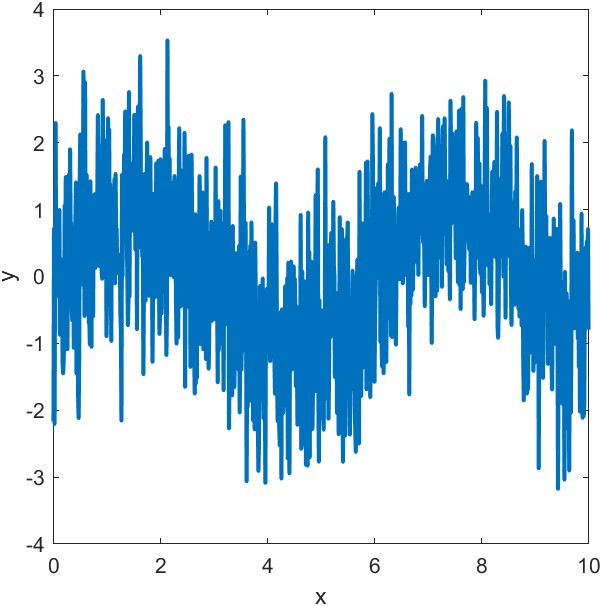}}\subfigure[]{\includegraphics[width=5.5cm]{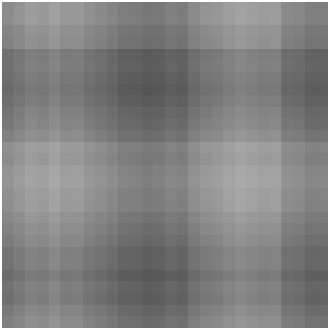}}
    \caption{(a) The curve of $y=\sin(\omega x)$ when $\omega=1.06$. (b) The 28 by 28 image that corresponds to the curve in (a). (c) The curve of $y=\sin(\omega x)+\sigma z$ when $\omega=1.06$ and $\sigma=.1$ where $z$ is a standard normal random variable. (d) The 28 by 28 image corresponds to the curve in (c). (e) The curve of $y=\sin(\omega x)+\sigma z$ when $\omega=1.06$ and $\sigma=1$ where $z$ is a standard normal random variable. (f) The 28 by 28 image that corresponds to the curve in (e).}
    \label{fig:figure4}
\end{figure}

\begin{figure}[H]
    \centering
    \subfigure[]{\includegraphics[width=5.6cm]{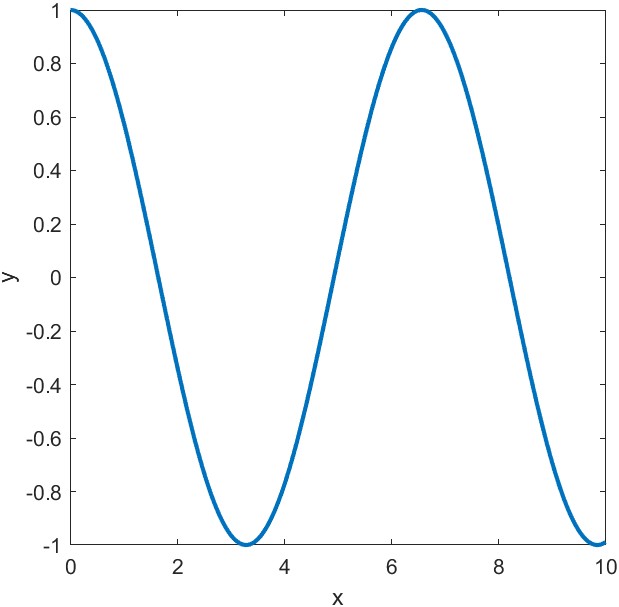}}\subfigure[]{\includegraphics[width=5.5cm]{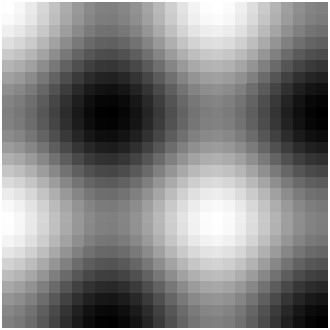}}
    \subfigure[]{\includegraphics[width=5.6cm]{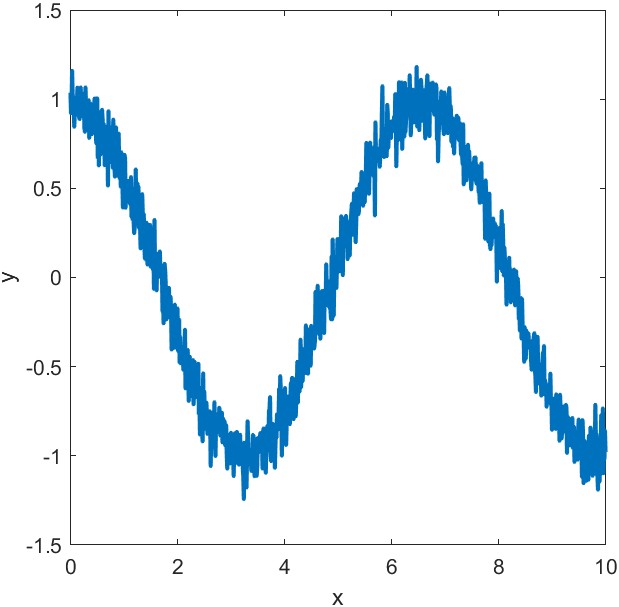}}\subfigure[]{\includegraphics[width=5.5cm]{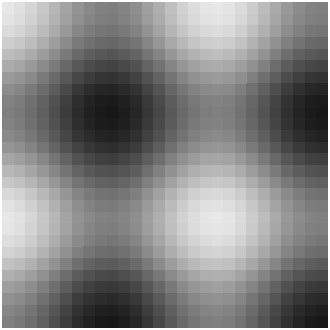}}
    \subfigure[]{\includegraphics[width=5.6cm]{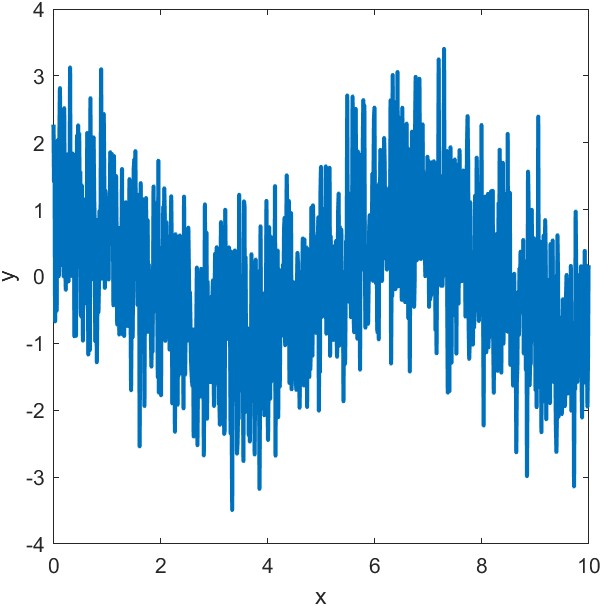}}\subfigure[]{\includegraphics[width=5.5cm]{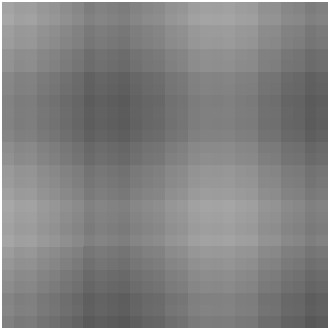}}
    \caption{(a) The curve of $y=\cos(\omega x)$ when $\omega=.96$. (b) The 28 by 28 image that corresponds to the curve in (a). (c) The curve of $y=\cos(\omega x)+\sigma z$ when $\omega=.96$ and $\sigma=.1$ where $z$ is standard normal random variable. (d) The 28 by 28 image corresponds to the curve in (c). (f) The curve of $y=\cos(\omega x)+\sigma z$ when $\omega=.96$ and $\sigma=1$ where $z$ is a standard normal random variable. (e) The 28 by 28 image that corresponds to the curve in (e).}
    \label{fig:figure5}
\end{figure}

Consider the functions $y=\sin(\omega x)$ and $y=\cos(\omega x)$ with the frequency parameter $\omega$. See some example curves in Figures \ref{fig:figure4} and \ref{fig:figure5} for $y=\sin(1.06x)$ and $y=\cos(0.96x)$, respectively. 
For training, validation, and testing data, the parameter $\omega$ is randomly uniformly sampled from the interval $[0,3]$ for $1000$, $100$, and $100$ times, respectively.

\

\begin{figure}[H]
    \centering
    \subfigure[]{\includegraphics[width=7cm]{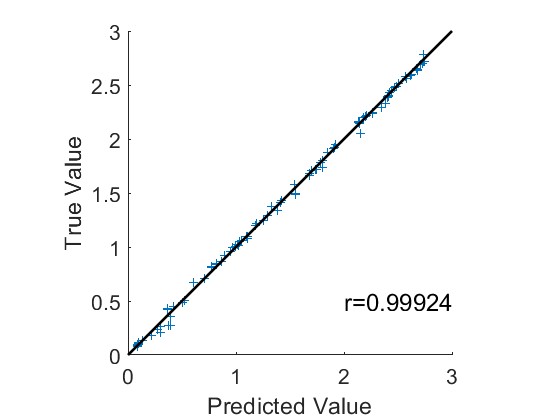}}\subfigure[]{\includegraphics[width=7cm]{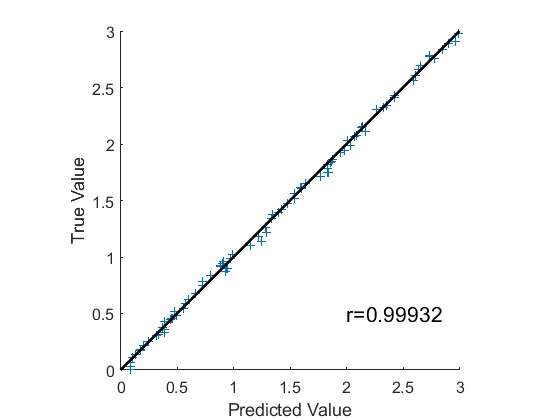}}
    \subfigure[]{\includegraphics[width=7cm]{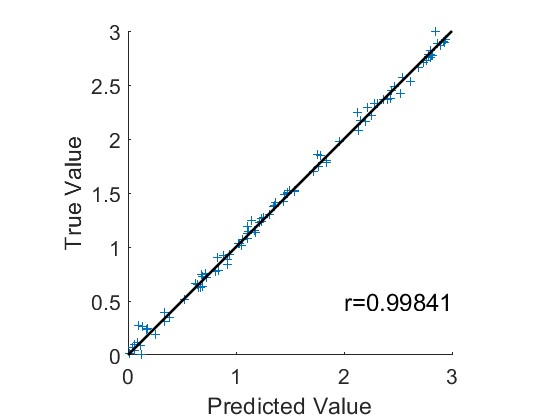}}\subfigure[]{\includegraphics[width=7cm]{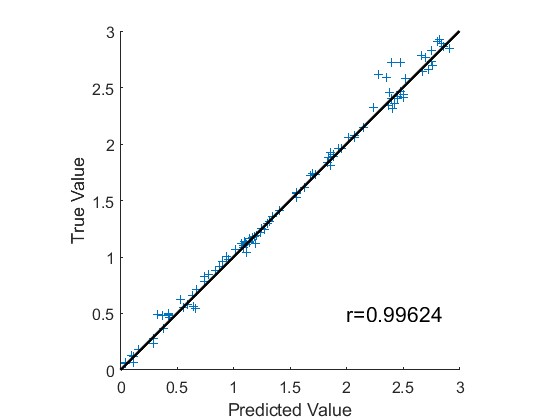}}
    \subfigure[]{\includegraphics[width=7cm]{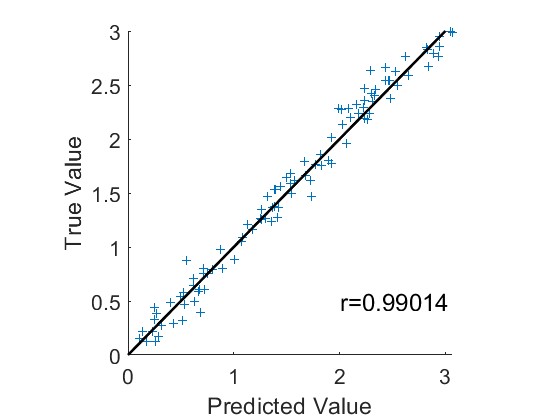}}\subfigure[]{\includegraphics[width=7cm]{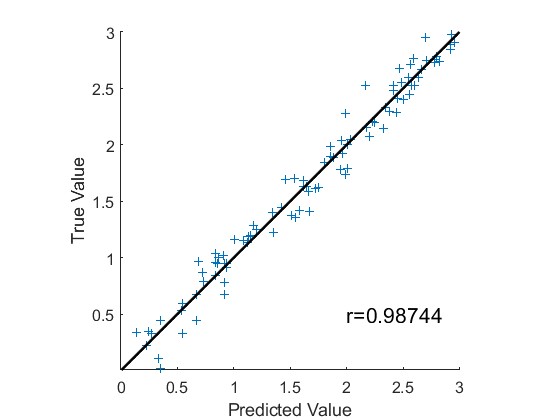}}
    \caption{Results of the sine function regression using the test dataset without noise (a), with noise of magnitude $\sigma=.1$ (c), and with noise of magnitude $\sigma=1$ (e). Results of the cosine function regression using the test dataset without noise (b), with noise of magnitude $\sigma=.1$ (d), and with noise of magnitude $\sigma=1$ (f).}
    \label{fig:figure6}
\end{figure}

Figure \ref{fig:figure6} shows the predicted values versus the estimated value of the parameter of the sine and cosine functions that closely follow the diagonal line without an intercept and slope of one. Tables \ref{tab:sine-results} and \ref{tab:cosine-results} show a strong diagonal linear relationship between the true and predicted bandwidth values.

\begin{table}[ht]
\centering
\caption{Correlation Coefficient, Intercept, and Slope for Sine Data with P-values}
\begin{tabular}{|c|c|c|c|c|}
\hline
Case & Correlation Coefficient ($r$) & Intercept (p-value) & Slope (p-value) \\
\hline
Without Noise & $\approx 1$ & 0.0015 (0.8306) &  0.9954 (0.9121) \\
\hline
With Noise ($\sigma=0.1$) & $\approx 1$ & 0.0171 (0.0836) & 0.9911 (0.4942) \\
\hline
With Noise ($\sigma=1$) & $\approx 1$ & 0.0011 (0.9649) & 1.0108 (0.2765) \\
\hline
\end{tabular}
\label{tab:sine-results}
\end{table}

\begin{table}[ht]
\centering
\caption{Correlation Coefficient, Intercept, and Slope for Cosine Data with P-values}
\begin{tabular}{|c|c|c|c|c|}
\hline
Case & Correlation Coefficient ($r$) & Intercept (p-value) & Slope (p-value) \\
\hline
Without Noise & $\approx 1$ & -0.0059 (0.3546) & 0.9979 (0.5825) \\
\hline
With Noise ($\sigma=0.1$) & $\approx 1$ & 0.0143 (0.3497) & 1.0045 (0.6084) \\
\hline
With Noise ($\sigma=1$) & $\approx 1$ & 0.0639 (0.0305) & 0.9647 (0.0259) \\
\hline
\end{tabular}
\label{tab:cosine-results}
\end{table}

\subsubsection{Estimation of the Height and Width of Peaks of Functions} \label{HeightWidth}
\begin{figure}[H]
    \centering
    \subfigure[]{\includegraphics[width=5.6cm]{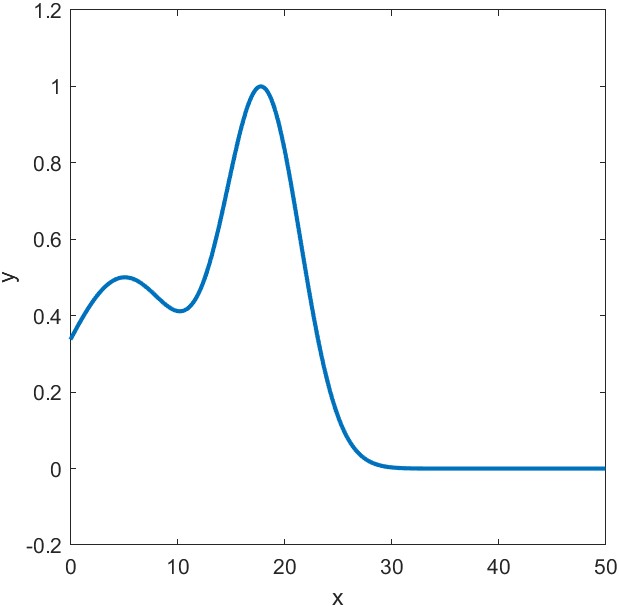}}\subfigure[]{\includegraphics[width=5.5cm]{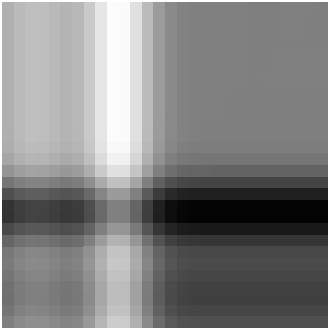}}
    \subfigure[]{\includegraphics[width=5.6cm]{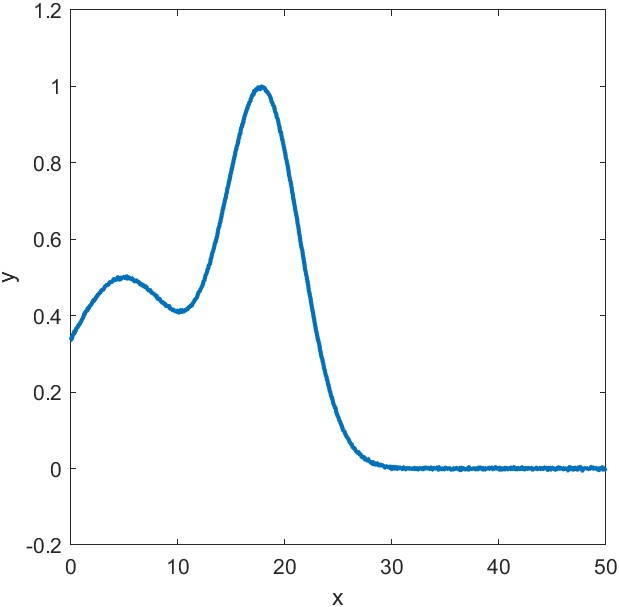}}\subfigure[]{\includegraphics[width=5.5cm]{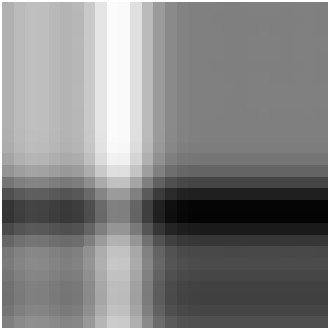}}    \subfigure[]{\includegraphics[width=5.5cm]{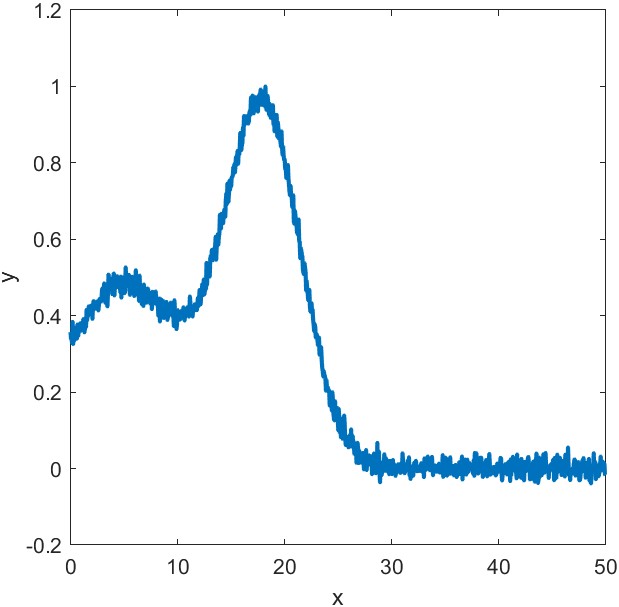}}\subfigure[]{\includegraphics[width=5.5cm]{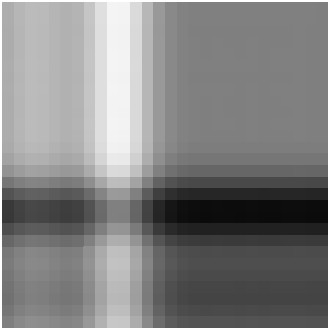}}
    \caption{(a) The curve is an example of a mixture of Gaussians $G(x)$. (b) The 28 by 28 image that corresponds to the curve in (a). (c) The curve of $G(x) +\sigma z$ when $\sigma=.1$ where $z$ is a standard normal random variable. (d) The 28 by 28 image that corresponds to the curve in (c). (e) The curve of $G(x)+\sigma z$ when  $\sigma=1$ where $z$ is a standard normal random variable. (f) The 28 by 28 image that corresponds to the curve in (e).}
    \label{fig:figure7}
\end{figure}

To produce a number of peaks with different heights and widths, we used a mixture of Gaussian curves, see Figures \ref{fig:figure7}. The maximum height of a curve is the height of the peak at the global maximum of the curve. The width of a curve is the horizontal distance of the contour located half of the prominence of that peak. It is important to note that the curves need to be normalized with the highest peak of all the generated curves to have grayscale values representative of all of the heights. Otherwise, the highest grayscale value of 1 will be assigned to the peak height of each curve and thus will not be able to train the model properly. On the other hand, with the width estimation and peak detection (see peak classification section), we use local normalization, since they do not necessarily depend on the height.

To generate the curves, we use a mixture of Gaussian kernels given by $y=G(x)=\sum^2_{k=1} H_k \exp({-(\frac{x-P_k}{W_k}})^2)$ for $x\in [0,50]$. Then the heights $H_k$ are randomly and uniformly selected from $[0,2200]$. The shape parameters $W_k$ are randomly generated using $W_k=\lfloor 50\cdot U_1+1 \rfloor $ where $U_1$ is a uniformly distributed random variable on [0,1]. Similarly, position parameters are randomly generated using $P_k=\lfloor 50\cdot U_2+1 \rfloor$ where $U_2$ is a uniformly distributed random variable over [0,1]. See Figure \ref{fig:figure7} for an example of a mixture of Gaussian kernels with and without random noise.

\begin{figure}[H]
    \centering
    \subfigure[]{\includegraphics[width=7cm]{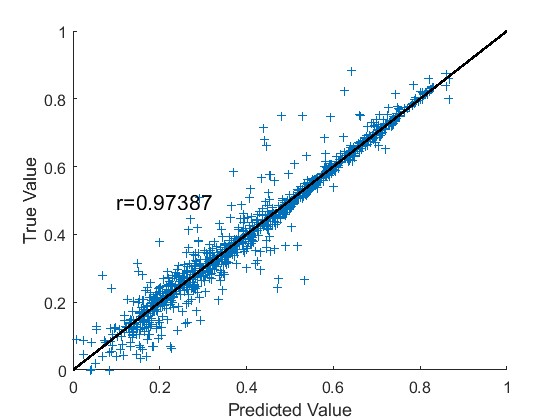}}  
    \subfigure[]{\includegraphics[width=7cm]{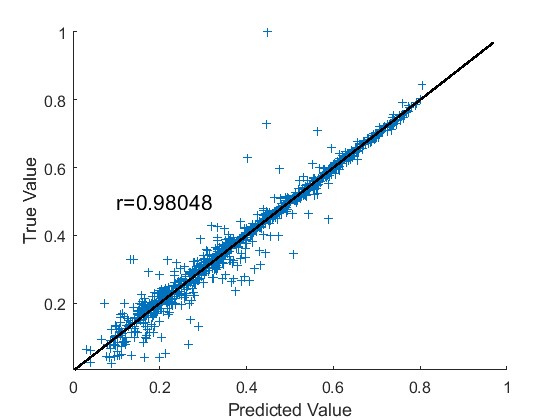}}
    \subfigure[]{\includegraphics[width=7cm]{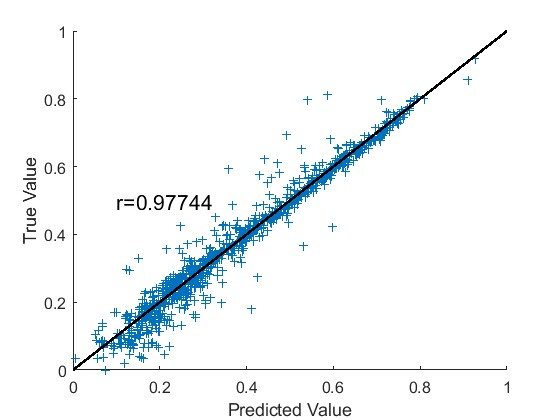}}
    \caption{Results of the maximum width regression using the test dataset without noise (a), with noise of magnitude $\sigma=.1$ (b), and with noise of magnitude $\sigma=1$ (c).}
    \label{fig:figure8}
\end{figure}

\begin{figure}[H]
    \centering
    \subfigure[]{\includegraphics[width=7cm]{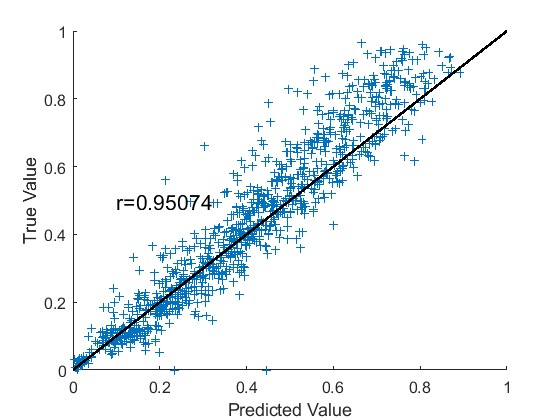}}  
    \subfigure[]{\includegraphics[width=7cm]{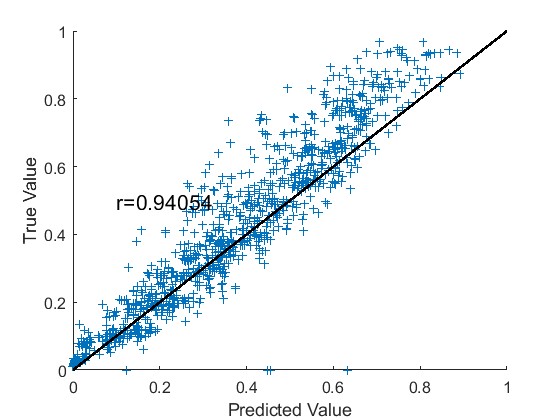}}
    \subfigure[]{\includegraphics[width=7cm]{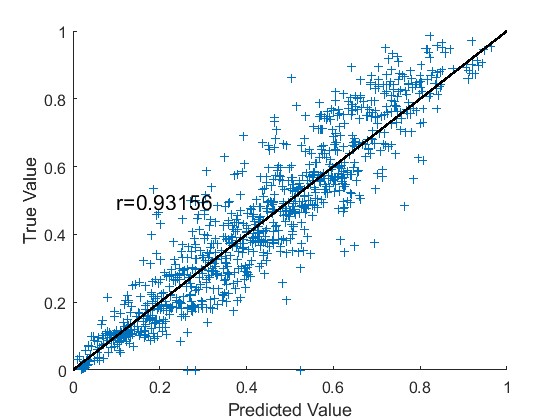}}
    \caption{Results of the maximum height regression using the test dataset without noise (a), with noise of magnitude $\sigma=.1$ (b), and with noise of magnitude $\sigma=1$ (c).}
    \label{fig:figure9}
\end{figure}

Figure \ref{fig:figure8} shows that the normalized width predictions closely follow the actual values. In this part, Table \ref{tab:width-estimation-results} shows a strong indication of accurate prediction of the width of the peaks of the curves. Table \ref{tab:height-estimation-results} supported by Figure \ref{fig:figure9}, however, shows that the CNN could not retrieve the actual values of height on average unless it has more noise.

\begin{table}[ht]
\centering
\caption{Correlation Coefficient, Intercept, and Slope for Width Estimation Data with P-values}
\begin{tabular}{|c|c|c|c|c|}
\hline
Case & Correlation Coefficient ($r$) & Intercept (p-value) & Slope (p-value) \\
\hline
Without Noise & 0.974 & -0.0017 (0.6247) & 1.0057 (0.4453) \\
\hline
With Noise ($\sigma=0.1$) & 0.980 & 0.0022 (0.4370) & 0.9964 (0.5736) \\
\hline
With Noise ($\sigma=1$) & 0.977 & -0.0071 (0.0178) & 1.0138 (0.0463) \\
\hline
\end{tabular}
\label{tab:width-estimation-results}
\end{table}

\begin{table}[ht]
\centering
\caption{Correlation Coefficient, Intercept, and Slope for Height Estimation Data with P-values}
\begin{tabular}{|c|c|c|c|c|}
\hline
Case & Correlation Coefficient ($r$) & Intercept (p-value) & Slope (p-value) \\
\hline
Without Noise & 0.951 & -0.0169 (0.0013) & 1.1195 (0.0000) \\
\hline
With Noise ($\sigma=0.1$) & 0.941 & 0.0084 (0.1244) & 1.1044 (0.0000) \\
\hline
With Noise ($\sigma=1$) & 0.932 & -0.0057 (0.3473) & 1.0327 (0.0105) \\
\hline
\end{tabular}
\label{tab:height-estimation-results}
\end{table}

\subsection{Classification Problems}
In this subsection, we examine the capabilities of CNN in functional data classification. For this type of problem, the same CNN architecture is used except that after the drop-out layer, the fully connected layer has hidden nodes equal to the number of classes, followed by a softmax layer followed by a classification Layer.

\subsubsection{Increasing versus Decreasing Curves}
\begin{figure}[H]
    \centering
    \subfigure[]{\includegraphics[width=7.5cm]{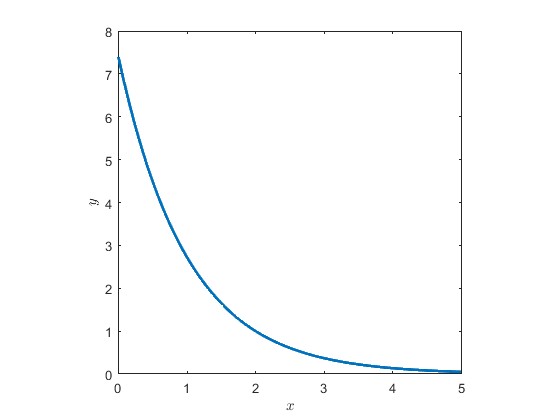}} 
    \subfigure[]{\includegraphics[width=5.25cm]{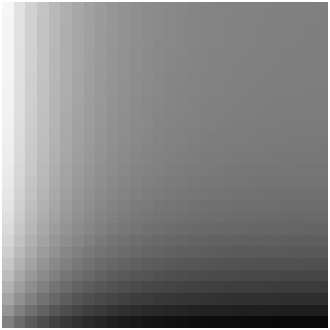}}
    \subfigure[]{\includegraphics[width=7.5cm]{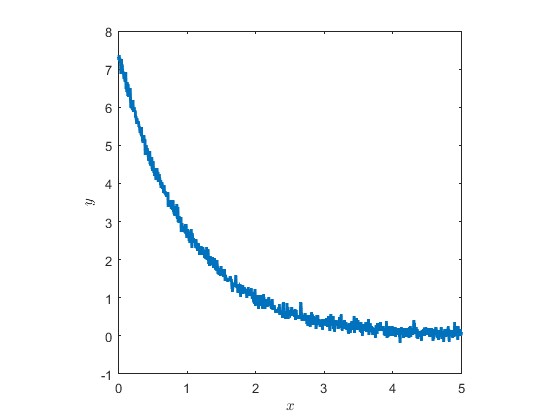}} 
    \subfigure[]{\includegraphics[width=5.25cm]{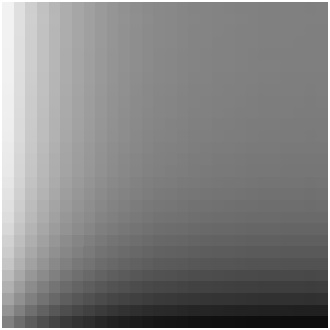}}
    \subfigure[]{\includegraphics[width=7.5cm]{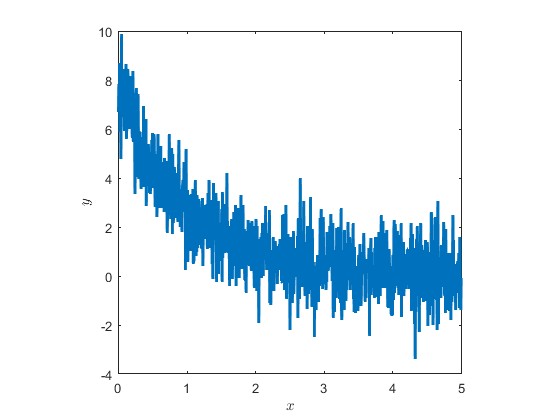}} 
    \subfigure[]{\includegraphics[width=5.25cm]{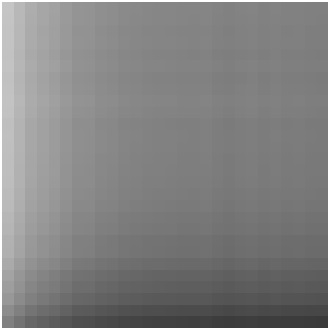}}
    \caption{(a)The curve of $y=e^{-(x-2)}$ . (b) The 28 by 28 image that corresponds to the curve in (a). (c) The curve of $e^{-(x-2)}+\sigma z$ when $\sigma=.1$ where $z$ is standard normal random variable. (d) The 28 by 28 image corresponds to the curve in (c). (e) The curve of $e^{-(x-2)}+\sigma z$ when $\sigma=1$ where $z$ is standard normal random variable. (f) The 28 by 28 image corresponds to the curve in (e).}
    \label{fig:figure10}.
\end{figure}

 We use CNN to classify the monotonicity of curves. Curves $y=e^{w_1(x-w_2)}$ are used to generate increasing or decreasing exponential curves for training, validation, and training datasets, see Figure \ref{fig:figure10}. We also use the following random variables $w_1=sign(U_1-.5)$ and $w_2=2U_2+2.5$ where $U_1$ and $U_2$ are uniformly distributed random variables on [0,1]. We found that the accuracy of the classification of the functional test data is 100\%.

\subsubsection{Convex versus Concave Curves}
\begin{figure}[H]
    \centering
    \subfigure[]{\includegraphics[width=7.5cm]{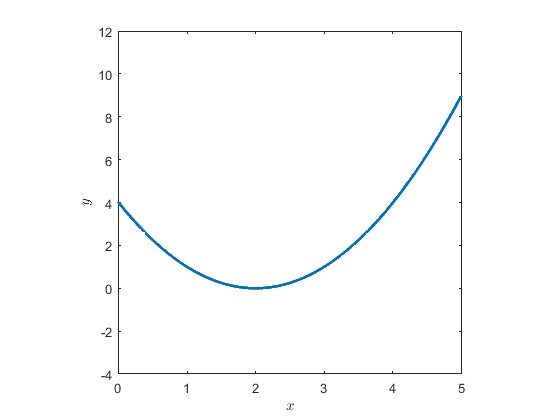}} 
    \subfigure[]{\includegraphics[width=5.25cm]{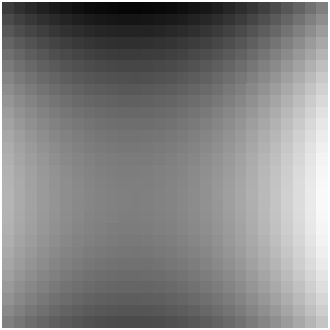}}
    \subfigure[]{\includegraphics[width=7.5cm]{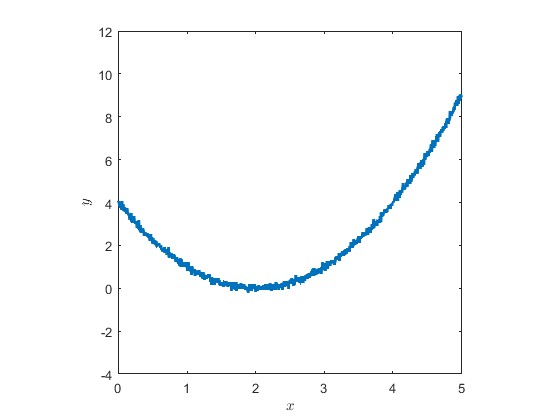}} 
    \subfigure[]{\includegraphics[width=5.25cm]{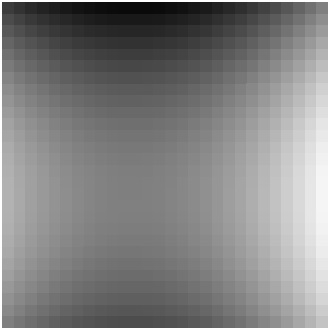}}
    \subfigure[]{\includegraphics[width=7.5cm]{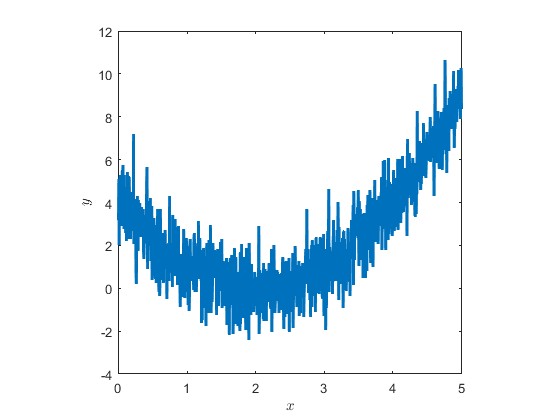}} 
    \subfigure[]{\includegraphics[width=5.25cm]{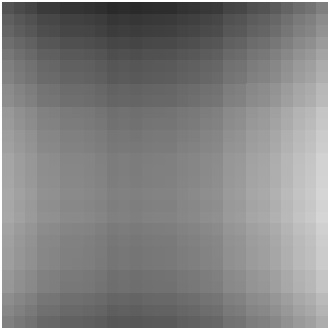}}
    \caption{(a)The curve of $y=(x-2)^2$ . (b) The 28 by 28 image that corresponds to the curve in (a). (c) The curve of $(x-2)^2+\sigma z$ when  $\sigma=.1$ where $z$ is a standard normal random variable. (d) The 28 by 28 image corresponds to the curve in (c). (e) The curve of $(x-2)^2+\sigma z$ when $\sigma=1$ where $z$ is a standard normal random variable. (f) The 28 by 28 image corresponds to the curve in (e).}
    \label{fig:figure11}
\end{figure}

We also examined the ability of CNN to classify the curvature of the curves as convex or concave. Curves $y=w_1(x-w_2)^2$ are used to generate convex and concave curves for training, validation, and training datasets; see Figure \ref{fig:figure11}. We also use the following random variables $w_1=sign(U_1-.5)$ and $w_2=2U_2+2.5$ where $U_1$ and $U_2$ are uniformly distributed random variables on [0,1]. We found that the accuracy of the classification of the functional test data is 100\%.

\subsubsection{Exponential versus Algebraic Growth Curves}\label{expvalg}
\begin{figure}[H]
    \centering
    \subfigure[]{\includegraphics[width=7cm]{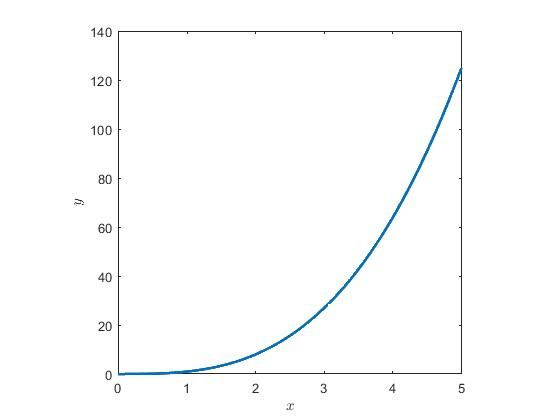}} 
    \subfigure[]{\includegraphics[width=5cm]{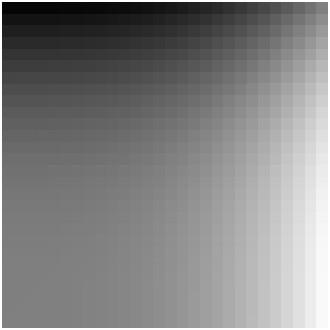}}
    \subfigure[]{\includegraphics[width=7cm]{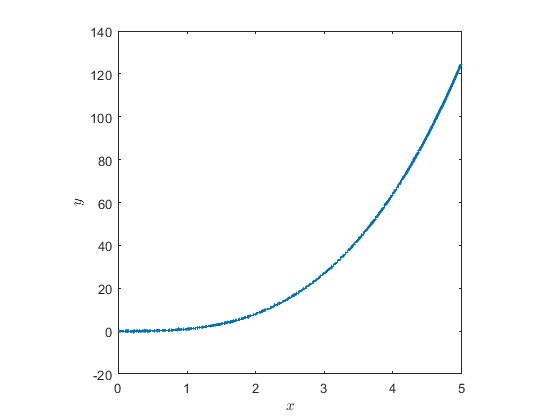}} 
    \subfigure[]{\includegraphics[width=5cm]{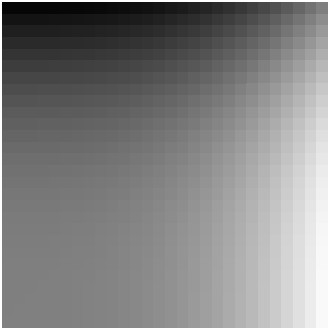}}
    \subfigure[]{\includegraphics[width=7cm]{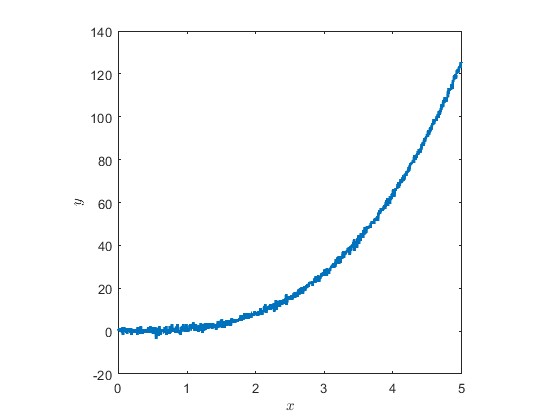}} 
    \subfigure[]{\includegraphics[width=5cm]{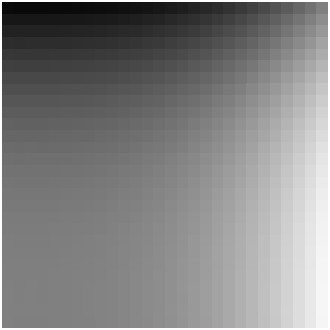}}
    \caption{(a)The curve of an algebraic growth represented by $y=x^3$. (b) The 28 by 28 image that corresponds to the curve in (a). (c) The curve of $y=x^3+\sigma z$ when  $\sigma=.1$ where $z$ is a standard normal random variable. (d) The 28 by 28 image corresponds to the curve in (c). (e) The curve of $y=x^3+\sigma z$ when $\sigma=1$ where $z$ is a standard normal random variable. (f) The 28 by 28 image corresponds to the curve in (e).}
    \label{fig:figure12}
\end{figure}

We examined the capabilities of CNN in classifying curve growth as exponential (in the form of $y=e^{cx}$ for $c>0$) or algebraic (in the form of $y=x^c$ for $c>0$), see Figure \ref{fig:figure12}. Training, validation, and training curves are generated using a random $c=3U+1$ where $U$ is a uniformly distributed random variable over [0,1]. We found that the classification accuracy of the functional test data is 100\%.

\subsubsection{Finding the Number of Peaks of Curves}
The same CNN architecture that is used to estimate the magnitude of the maximum height and the width of the peaks is used for classification with the number of peaks as mentioned in Subsection \ref{HeightWidth}. The curves are generated with the same scheme as mentioned in Subsection \ref{HeightWidth}. In this case, we use a local normalization; that is, the highest grayscale value of that particular curve is 1, and the lowest is 0. This helps to identify the peaks; since we are not interested in the height, the grayscale values do not need to be proportional to the height. A zero number of peaks is possible if the randomly generated position is outside the interval [0,50]. We found that the classification accuracy of the functional test data is 98. 0\% for noise-free data, 97. 2\% for noisy data with magnitude $\sigma=.1$ and 96. 2\% for noisy data with magnitude $\sigma=1$.

\section{Applications to Dynamical Systems}
In this section, we apply a CNN to dynamical systems. First, we show how CNN can estimate the Lyapunov exponent from one curve of the three variables solving the Lorenz system. Second, we show that CNN can be used to estimate transmission rates from epidemic curves. Third, we use CNN to test the similarity and dissimilarity of drug dissolution profiles.

\subsection{Estimating Lyapunov Exponent}
The study of human motion has been the focus of interest in the medical field to determine which exercise and range of motion are stable. Different perspectives have been performed to study such stability. Although \cite{stergiou2011human} pointed out the link between movement variability and stability. \cite{wilson2008coordination} emphasized that variability should not be confused with instability, as it can be observed in healthy and unhealthy subjects. \cite{dingwell2000nonlinear} helped develop the standard procedure for analyzing stability using Lyapunov exponents estimated by the Rosenstein method \cite{rosenstein1993practical}. Lyapunov exponents are the average exponential divergence of the orbits between nearest trajectories, which are called nearest neighbors. The more unstable the system, the higher the value of the Lyapunov exponent. An alternative parameter is the mean of the absolute value of the Floquet multipliers. \cite{hurmuzlu1994measurement} is the first to extensively use Floquet multipliers for the same purpose.  Floquet multipliers are the eigenvalues of the Jacobian matrix that measure the separation between orbits of the system \cite{dingwell2007differences} that could be calculated using Poincare maps. If the mean of the magnitudes of the eigenvalues is less than 1, the orbits are considered stable. 

\begin{figure}[H]
    \centering
    \subfigure[]{\includegraphics[width=7cm]{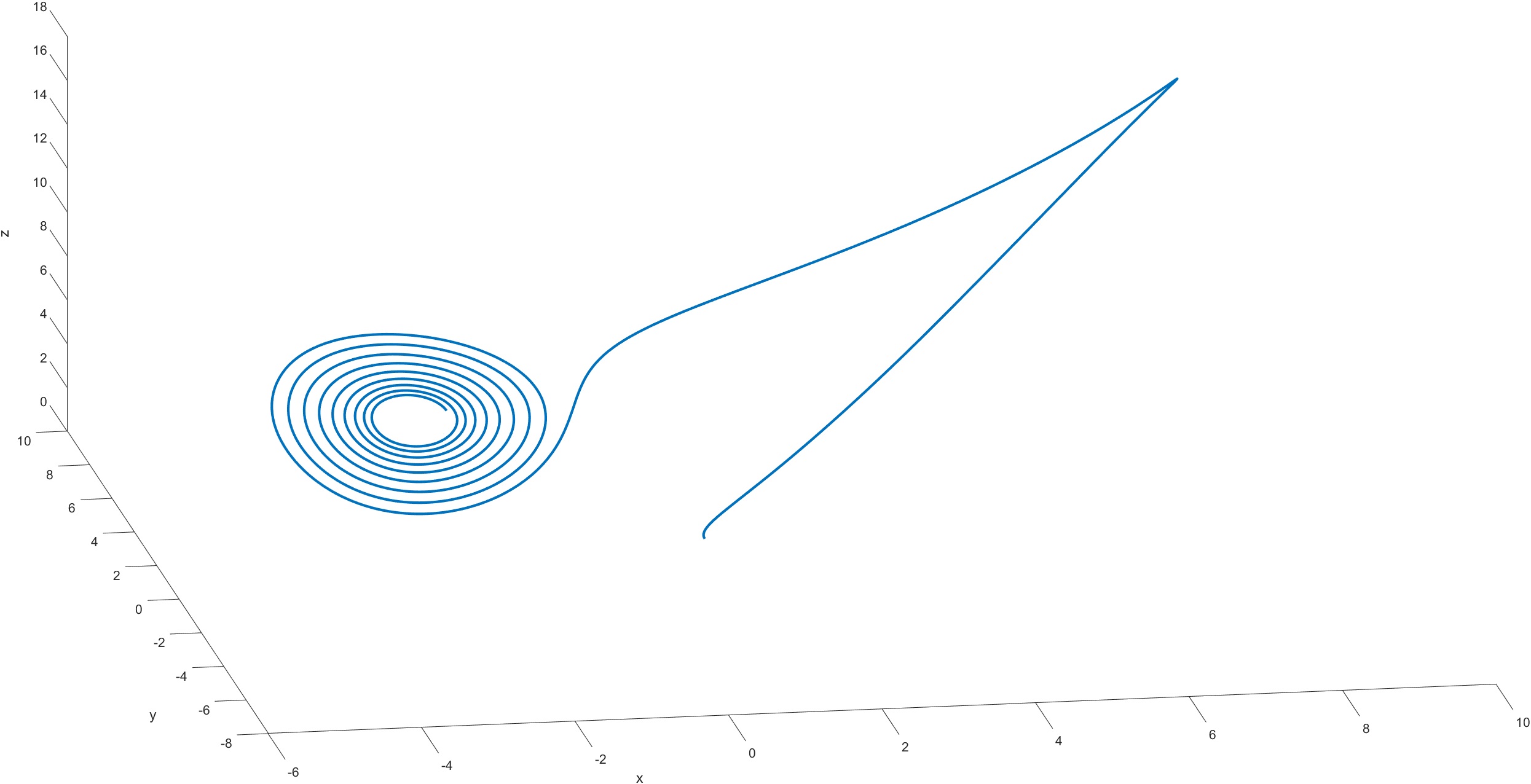}}
    \subfigure[]{\includegraphics[width=7cm]{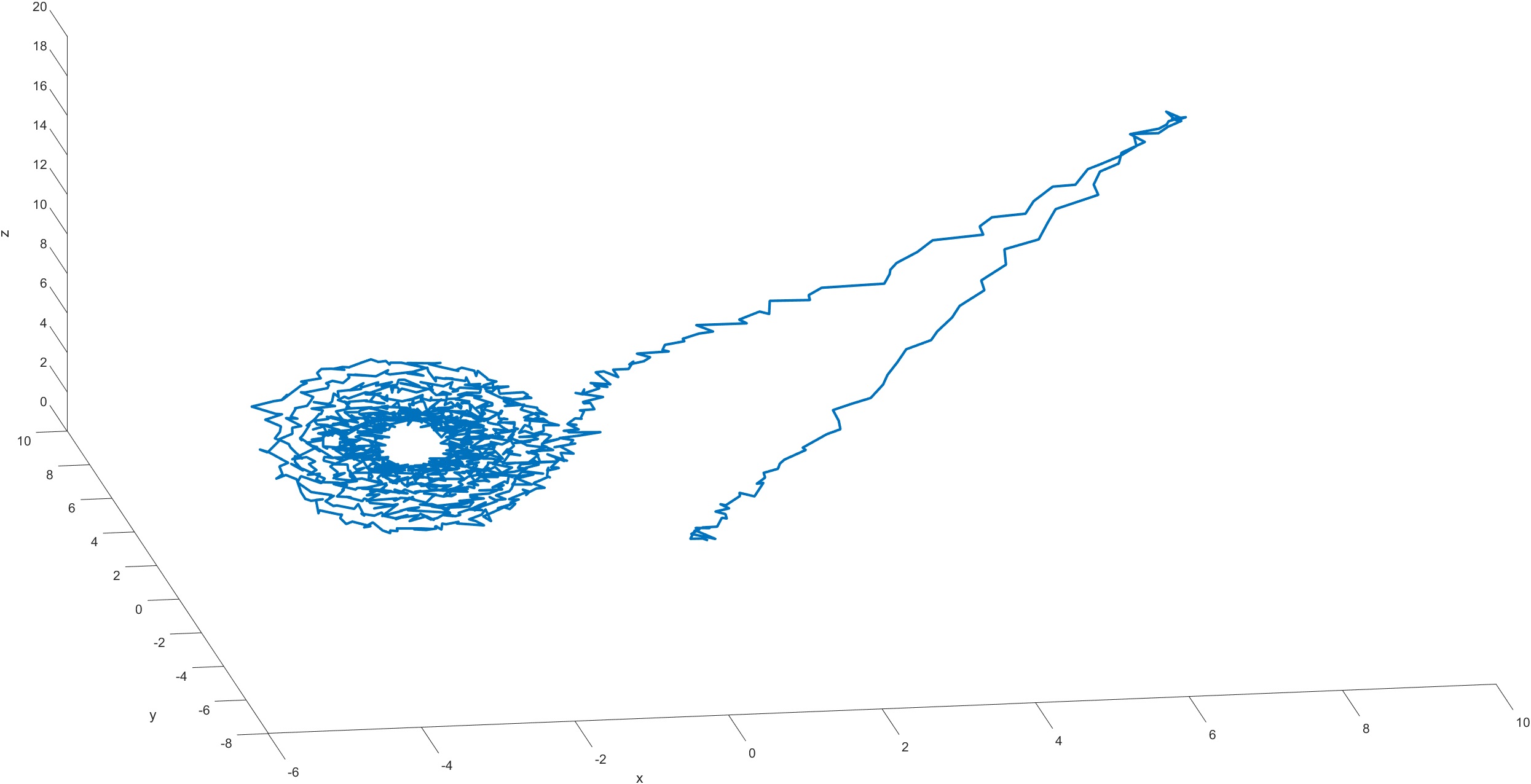}}
    \subfigure[]{\includegraphics[width=7cm]{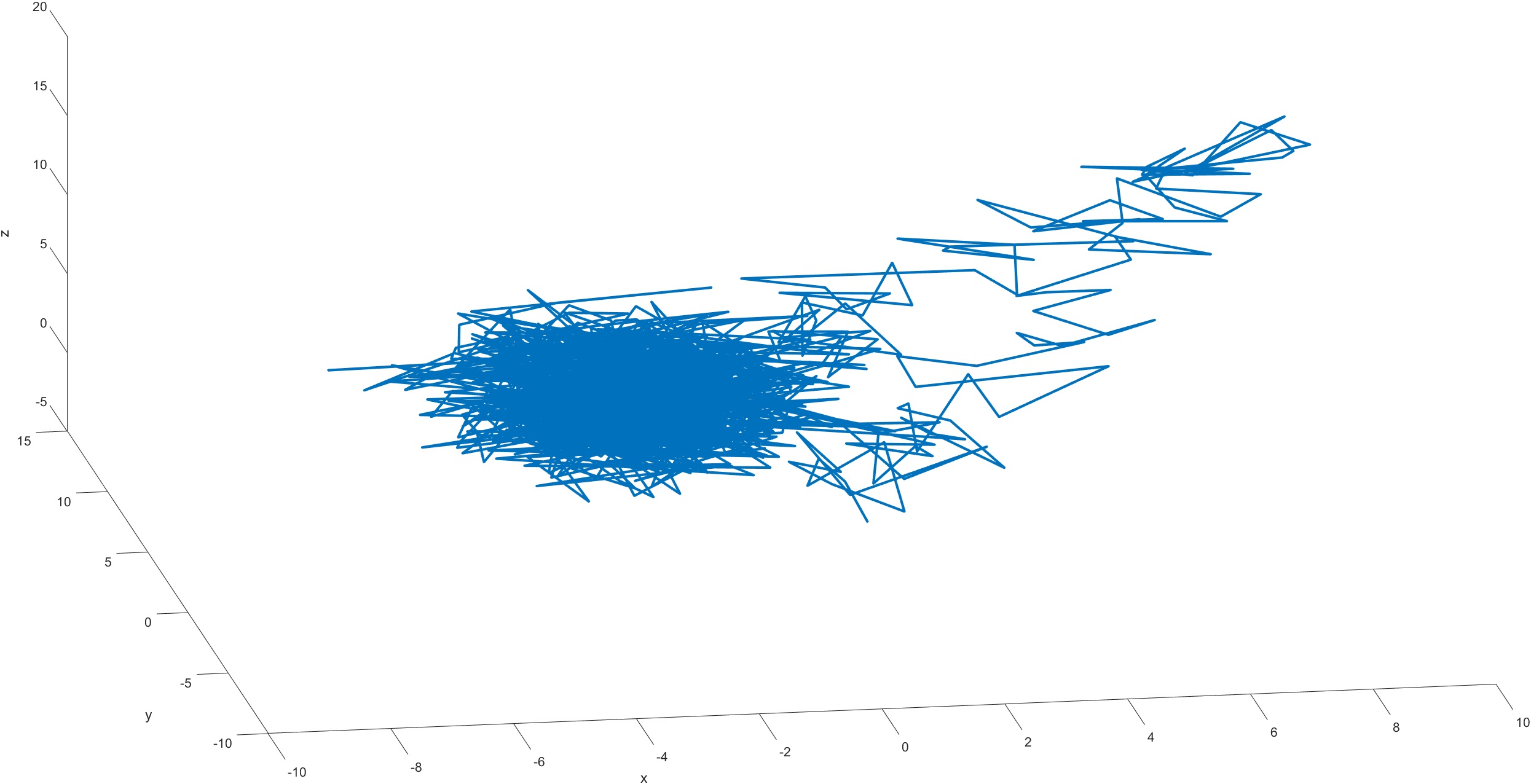}}
    \caption{ An example of Lorenz attractor at $\alpha=2.8029$, $\beta=1.1114$, and $\rho=11.9620$ without noise (a), with noise of magnitude $\sigma=.1$ (b), and with noise of magnitude $\sigma=1$ (c).
    }
    \label{fig:figure13}
\end{figure}

To examine our method in estimating the Lyapunov exponent we use the prototype of chaotic systems, Lorenz systems with parameters $\alpha, \beta, \rho$ defined as the system of ordinary differential equations:
\begin{eqnarray*}
\frac{dx}{dt} &=& \alpha (y - x) \\
\frac{dy}{dt} &=& x(\rho - z) - y \\
\frac{dz}{dt} &=& xy - \beta z
\end{eqnarray*}
Initial values are $x(0)=1$, $y(0)=1$, and $z(0)=1$. Figure \ref{fig:figure13} shows an example image of the attractor with $\alpha=2.8029$, $\beta=1.1114$, and $\rho=11.9620$.

\begin{figure}[H]
    \centering
    \subfigure[]{\includegraphics[width=5.5cm]{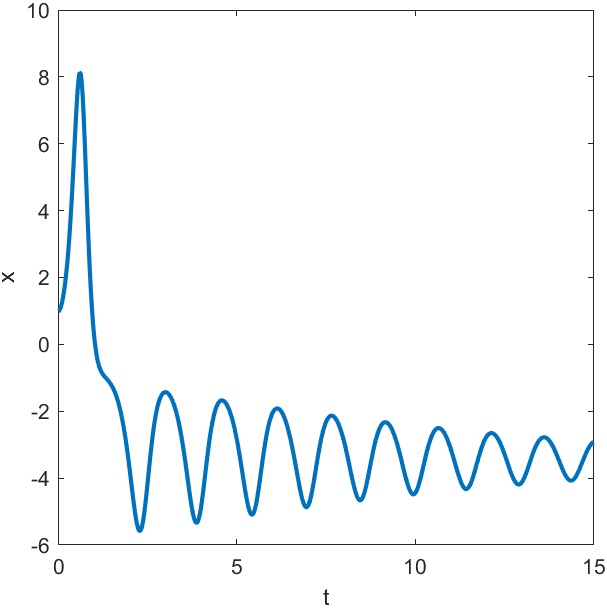}} 
    \subfigure[]{\includegraphics[width=5.5cm]{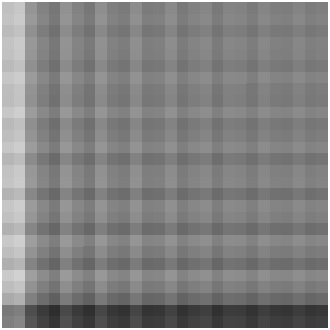}}
    \subfigure[]{\includegraphics[width=5.5cm]{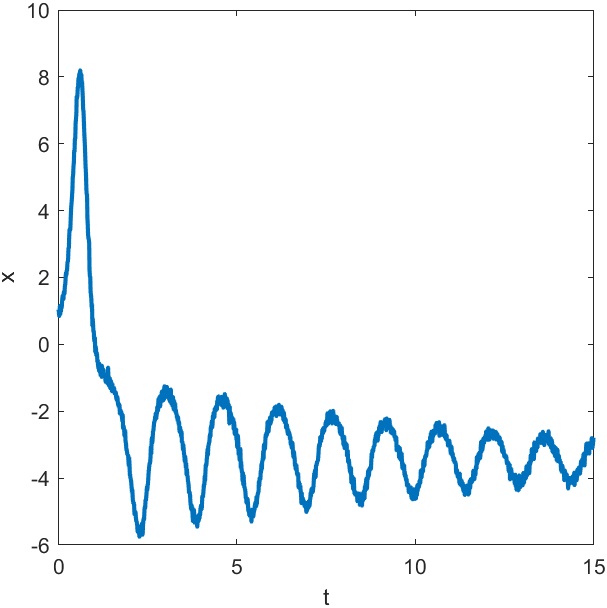}} 
    \subfigure[]{\includegraphics[width=5.5cm]{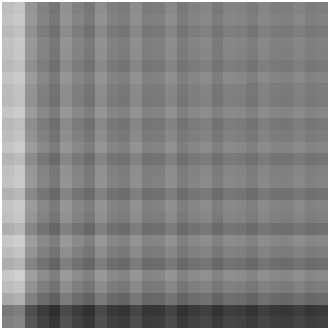}}
    \subfigure[]{\includegraphics[width=5.5cm]{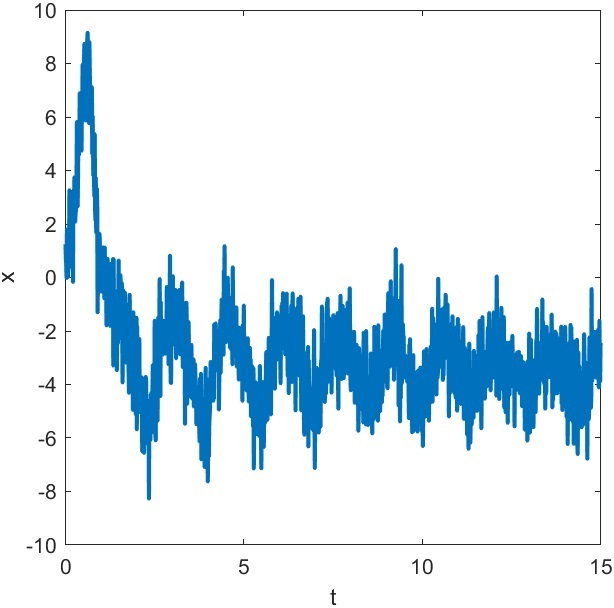}} 
    \subfigure[]{\includegraphics[width=5.5cm]{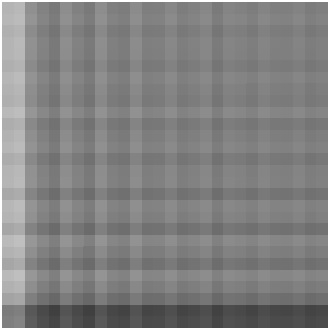}}
    \caption{(a) The curve of component $x$ of the Lorenz attractor. (b) The 28 by 28 image that corresponds to the curve in (a). (c) The curve of $x+\sigma w$ when $\sigma=.1$ where $w$ is a standard normal random variable. (d) The 28 by 28 image corresponds to the curve in (c). (e) The curve of $x+\sigma w$ when $\sigma=1$ where $w$ is a standard normal random variable. (f) The 28 by 28 image corresponds to the curve in (e).}
    \label{fig:figure14}
\end{figure}

We randomly simulate the values $\alpha=10\,U_1$, $\beta=8/3 \, U_2$,  and $\rho=20\, U_3$,  where $U_i$ are independent uniformly distributed random variables in $[0,1]$ for $i=1,2,3$. We ran the Lorenz system simulation over the time interval $[0,1]$ using the Runge-Kutta hybrid order 4 and 5 numerical method in MATLAB. We use component $x$ only for the training, validation, and testing of the CNN, see Figure \ref{fig:figure14}.

\begin{figure}[H]
    \centering
    \subfigure[]{\includegraphics[width=7cm]{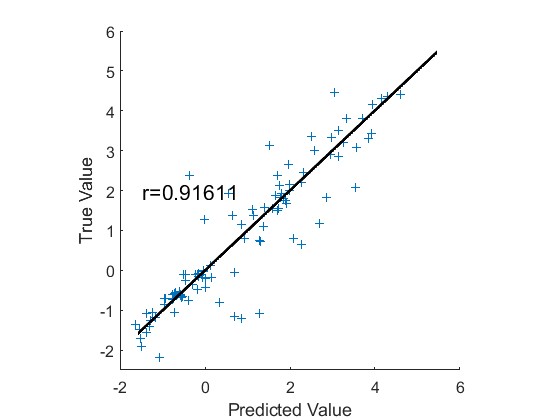}}  
    \subfigure[]{\includegraphics[width=7cm]{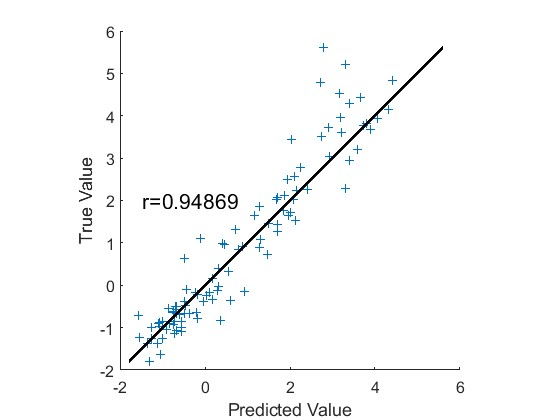}}
    \subfigure[]{\includegraphics[width=7cm]{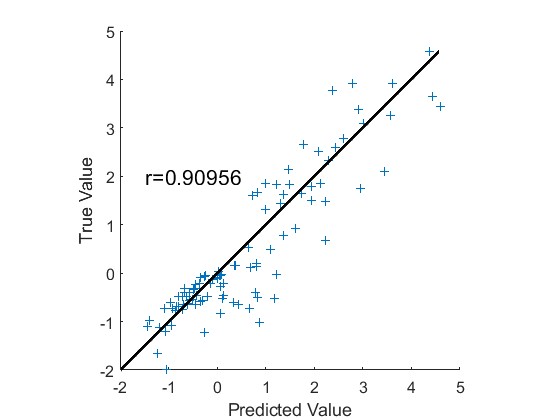}}
    \caption{Results of Lyapunov exponent estimation using the test dataset without noise (a), with noise of magnitude $\sigma=.1$ (b), and with noise of magnitude $\sigma=1$ (c).}
    \label{fig:figure15}
\end{figure}

Figure \ref{fig:figure15} shows the predicted values versus the estimated value of the Lyapunov exponents that closely followed the diagonal line without an intercept and a slope of one. Table \ref{tab:Lyapunov-results} shows a strong diagonal linear relationship between the true and predicted values of the Lyapunov exponents.

\begin{table}[ht]
\centering
\caption{Correlation Coefficient, Intercept, and Slope for Noise-Free Testing Data with P-values}
\begin{tabular}{|c|c|c|c|c|}
\hline
Case & Correlation Coefficient ($r$) & Intercept (p-value) & Slope (p-value) \\
\hline
Without Noise & 0.916 & -0.0146 (0.8550) & 0.9600 (0.3477) \\
\hline
With Noise ($\sigma=0.1$) & 0.949 & 0.0109 (0.8734) & 1.0908 (0.0152) \\
\hline
With Noise ($\sigma=1$) & 0.910 & -0.1140 (0.0900) & 0.9369 (0.1476) \\
\hline
\end{tabular}
\label{tab:Lyapunov-results}
\end{table}

\begin{figure}[H]
    \centering
    \subfigure[]{\includegraphics[width=7cm]{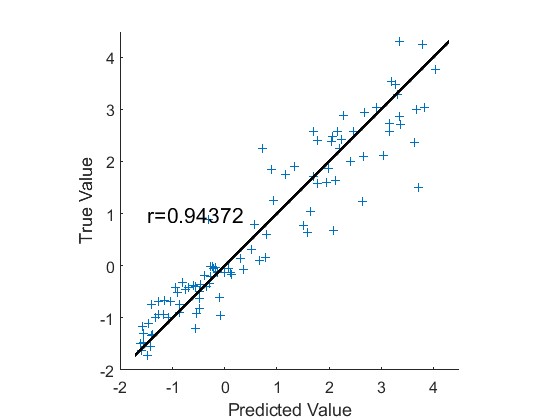}}  
    \subfigure[]{\includegraphics[width=7cm]{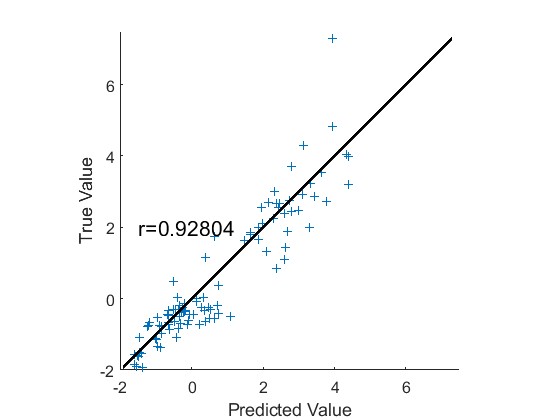}}
    \subfigure[]{\includegraphics[width=7cm]{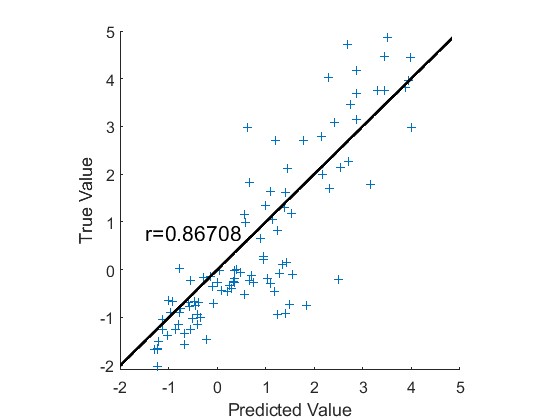}}
    \caption{Results of Lyapunov exponent estimation when the CNN is trained only using the noise-free data for the test dataset without noise (a), with noise of magnitude $\sigma=.1$ (b), and with noise of magnitude $\sigma=1$ (c).}
    \label{fig:figure16}
\end{figure}

Also, we tested the capability that a CNN model trained with noise-free data in estimating the Lyapunov exponent in noisy data. Figure \ref{fig:figure16} and Table \ref{tab:new-Lyapunov-results} show relatively good results. 

\begin{table}[ht]
\centering
\caption{Correlation Coefficient, Intercept, and Slope for Lyapunov Testing Data with P-values when CNN is Trained with Noise-Free Data.}
\begin{tabular}{|c|c|c|c|c|}
\hline
Case & Correlation Coefficient ($r$) & Intercept (p-value) & Slope (p-value) \\
\hline
Without Noise & 0.944 & 0.0567 (0.3275) & 0.8856 (0.0004) \\
\hline
With Noise ($\sigma=0.1$) & 0.928 & -0.1326 (0.0733) & 0.9871 (0.7472) \\
\hline
With Noise ($\sigma=1$) & 0.867 & -0.3390 (0.0015) & 1.0815 (0.1970) \\
\hline
\end{tabular}
\label{tab:new-Lyapunov-results}
\end{table}

An endurance test was also performed using 10000 test curves to estimate the Lyapunov exponent and found that the CNN is approximately 600 times faster (2.7628 seconds) compared to the MATLAB Rosenstein method (1692.6 seconds).

\subsection{Estimating the Transmission Rates from Epidemic Curves}
Estimation of transmission rates and exponential growth curve rates (see Subsection \ref{exp}) are important for emerging epidemics \cite{boonpatcharanon2022estimating} such as at the beginning of COVID-19 \cite{Tuite2020}. Some epidemics grow algebraically and not exponentially \cite{Chowell2015,Kolebaje2022} and so it is also helpful to discern them through classification; see Subsection \ref{expvalg}. One of the main goals in epidemiology is to estimate the basic reproduction number $R_0$ which is almost always proportional to the transmission rate $\beta$. If $R_0<1$, then the disease diminishes; otherwise, there is a chance that it will become endemic. 

We use a susceptible-infected-recovered (SIR) to produce epidemic curves with different transmission rates $\beta$ and estimate that parameter. The disease dynamics of a susceptible-infected-recovered (SIR) compartmental model follows the system of differential equations 
\begin{eqnarray*} \label{E2.1}
\frac{dS}{dt} &=& \mu - \beta \, S\,I-\mu \, S \\ 
\frac{dI}{dt}&=& \beta \, S\,I-(\mu+\gamma) \, I  \\ 
\frac{dR}{dt} &=& \gamma \, I-\mu \, R   
\end{eqnarray*}   
where $S$, $I$, and $R$ are the proportion of susceptible, infected, and recovered individuals in the population at time $t$, such that $S+I+R=1$. Initial values are $S(0)=.99$, $I(0)=.01$ and $R(0)=0$. The parameter $\mu$ is the per capita birth/death rate, $\beta$ is the transmission rate, and $\gamma$ is the recovery rate. The basic reproduction number in the above SIR model is given by $R_0=\beta/(\mu+\gamma)$. 

\begin{figure}[H]
    \centering
    \subfigure[]{\includegraphics[width=5cm]{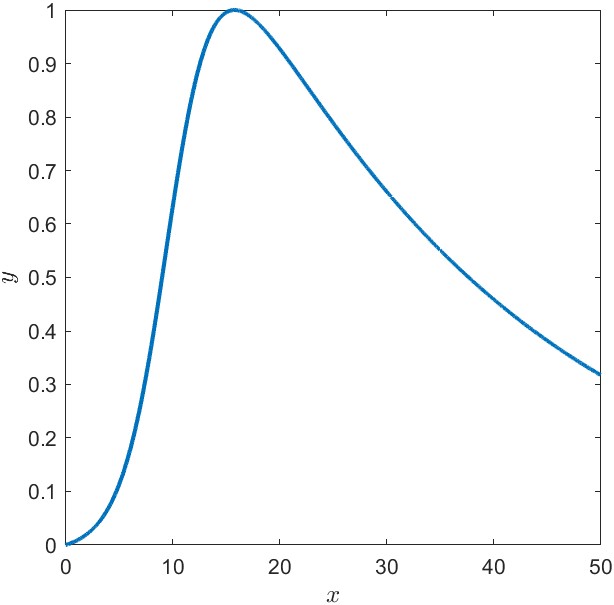}} 
    \subfigure[]{\includegraphics[width=5cm]{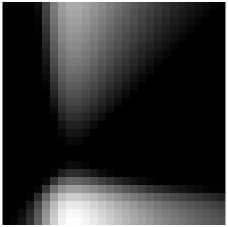}}
    \subfigure[]{\includegraphics[width=5cm]{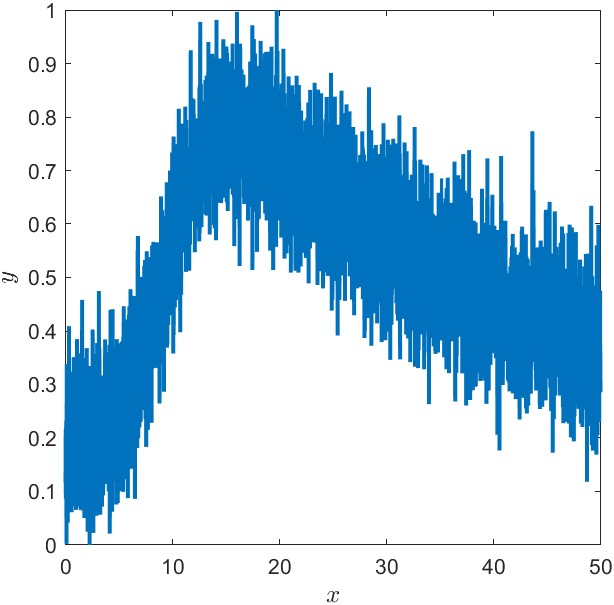}} 
    \subfigure[]{\includegraphics[width=5cm]{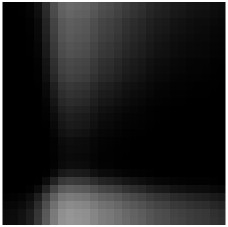}}
    \subfigure[]{\includegraphics[width=5cm]{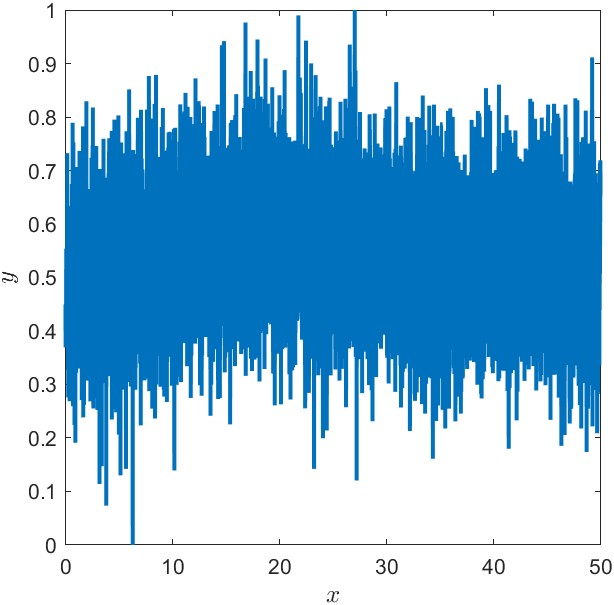}} 
    \subfigure[]{\includegraphics[width=5cm]{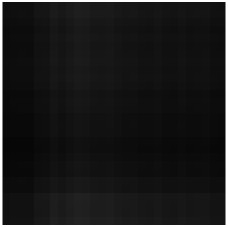}}
    \caption{(a) A simulation curve of $I$. (b) The 28 by 28 image that corresponds to the curve in (a). (c) The curve of $I+\sigma z$ when  $\sigma=.1$ where $z$ is a standard normal random variable. (d) The 28 by 28 image corresponds to the curve in (c). (f) The curve of $I+\sigma z$ when  $\sigma=1$ where $z$ is a standard normal random variable. (f) The 28 by 28 image corresponds to the curve in (e).}
    \label{fig:figure17}
\end{figure}

We assume values $\mu=1/(365*50)$ days$^{-1}$, $\gamma=1/28$ days$^{-1}$ and $\beta$ randomly selected from a uniform distribution over $(.01,1)$ to reflect a basic reproduction number in the range of $(.28,28)$. The SIR model simulations are run for $50$ days using the Runge-Kutta hybrid order 4 and 5 numerical method in MATLAB. See Figure \ref{fig:figure17} for a simulated epidemic curve $I$.

\begin{figure}[H]
    \centering
    \subfigure[]{\includegraphics[width=7cm]{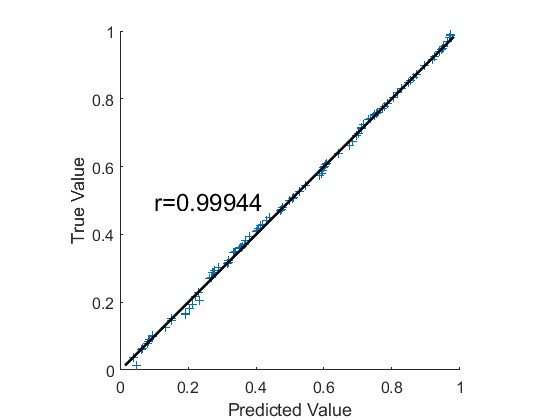}}  
    \subfigure[]{\includegraphics[width=7cm]{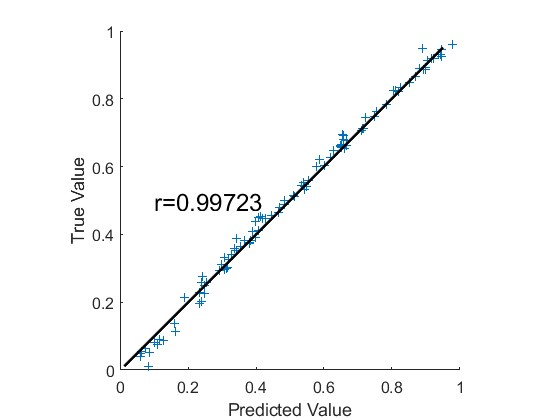}}
    \subfigure[]{\includegraphics[width=7cm]{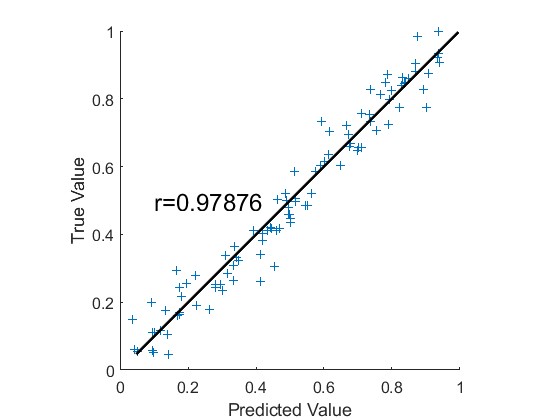}}
    \caption{Results of SIR regression using the test set. (a) No noise data.(b) noisy data with $\sigma=.1$. (c) noisy data with $\sigma=1$.}
    \label{fig:figure18}
\end{figure}

Figure \ref{fig:figure18} and Table \ref{tab:transmission-results} show strong prediction of transmission rates. 

\begin{table}[ht]
\centering
\caption{Correlation Coefficient, Intercept, and Slope for Transmission Rate Testing Data with P-values}
\begin{tabular}{|c|c|c|c|c|}
\hline
Case & Correlation Coefficient ($r$) & Intercept (p-value) & Slope (p-value) \\
\hline
Without Noise & 0.999 & -0.0023 (0.2573) & 1.0048 (0.1642) \\
\hline
With Noise ($\sigma=0.1$) & 0.997 & -0.0066 (0.1318) & 1.0177 (0.0229) \\
\hline
With Noise ($\sigma=1$) & 0.979 & 0.0005 (0.9686) & 0.9920 (0.7029) \\
\hline
\end{tabular}
\label{tab:transmission-results}
\end{table}

\subsection{Detecting Similarity of Drug Dissolution Profiles}
The problem of drug release or dissolution profiles is important for the pharmaceutical industry. Regulatory guidelines seek to advise coherent characteristics of drug dissolution prior to their approval; see, e.g. \cite{Vranic2002,Pourmohamad2023}. Many statistical approaches have been developed to test the similarity between dissolution curves or profiles, including cluster analysis, decision trees, and linear models; see, e.g. \cite{Costa2001,Maggio2008,Enachescu2010,Paixao2017,Abend2023,Pourmohamad2023}. That also includes nonparametric measures such as the two measures adopted by the US Food and Drug Administration (FDA) and the European Medicines Agency (EMA), $$ f_1 = \frac{\sum_{i=1}^n |R_i-S_i|}{\sum_{i=1}^n R_i} \times 100\%$$ and 
$$f_2=50 \log_{10} \left( \left[  1+\frac1n \sum_{i=1}^n (R_i-S_i)^2\right]^{-0.5} \times 100 \right)$$ which are widely used to detect dissimilarity and similarity, respectively, between two curves $\{(t_i,R_i):i=1,2,\ldots,n\}$ 
and
$\{(t_i,S_i):i=1,2,\ldots,n\}$. If $f_1$ is between 0 and 15 and $f_2$ is between 50 and 100, then the two curves are considered similar; see, for example,  \cite{Costa2001,Pourmohamad2023} for a complete set of models and measures, as well as FDA \& EMA guidelines. 

\begin{figure}[H]
    \centering
    \subfigure[]{\includegraphics[width=7cm]{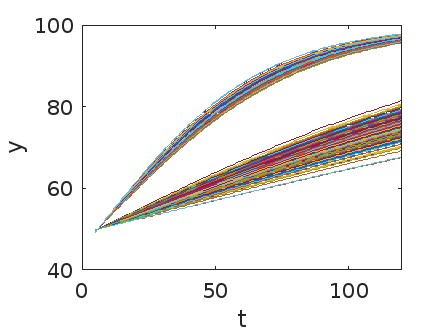}} 
    \subfigure[]{\includegraphics[width=7cm]{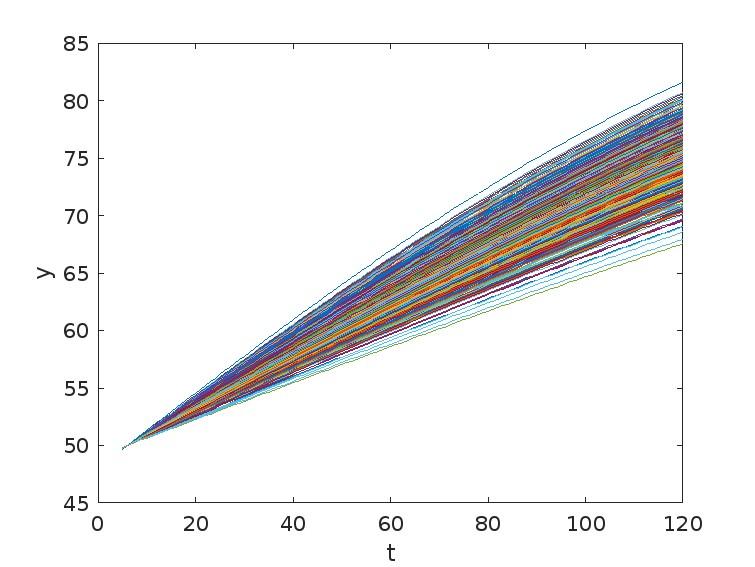}}   
    \subfigure[]{\includegraphics[width=7cm]{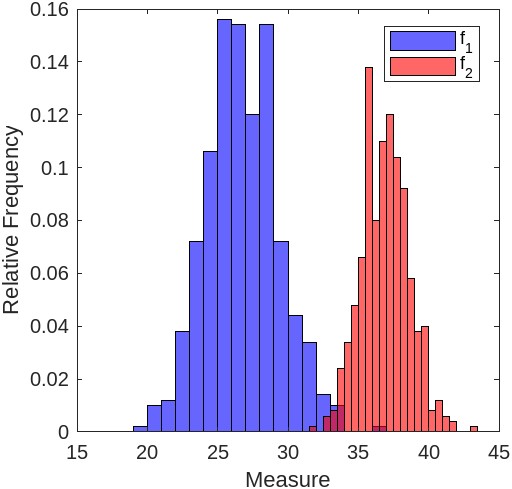}} 
    \subfigure[]{\includegraphics[width=7cm]{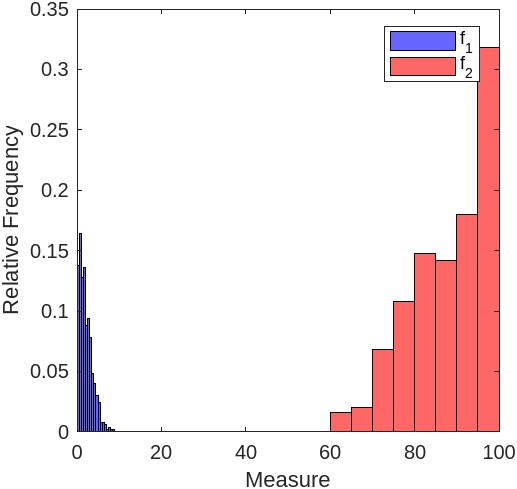}}
    \subfigure[]{\includegraphics[width=7cm]{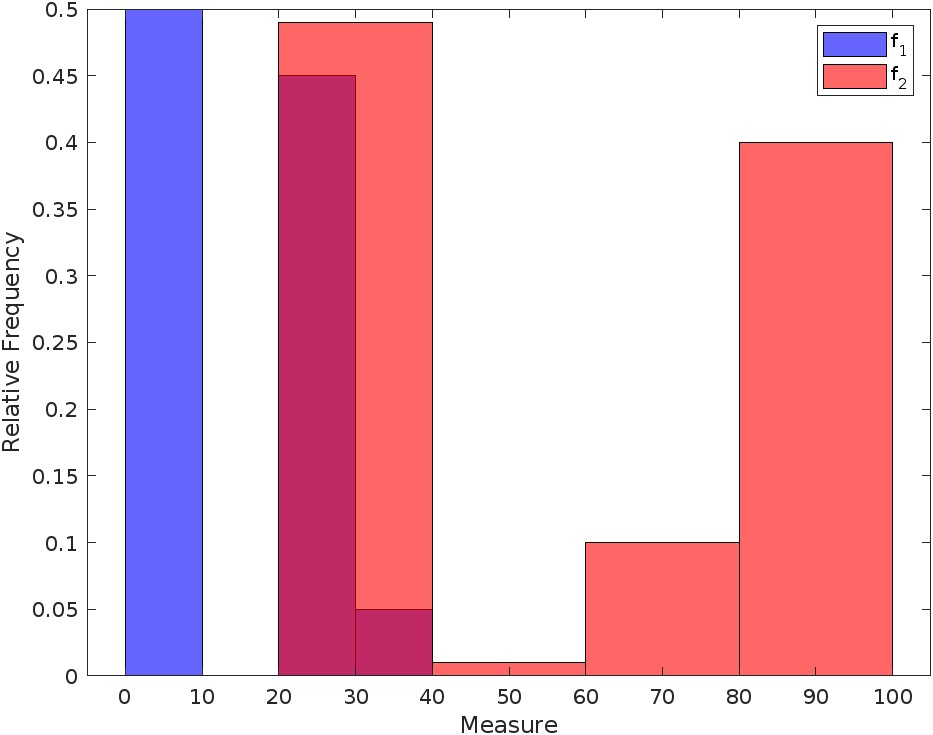}} 
    \subfigure[]{\includegraphics[width=7cm]{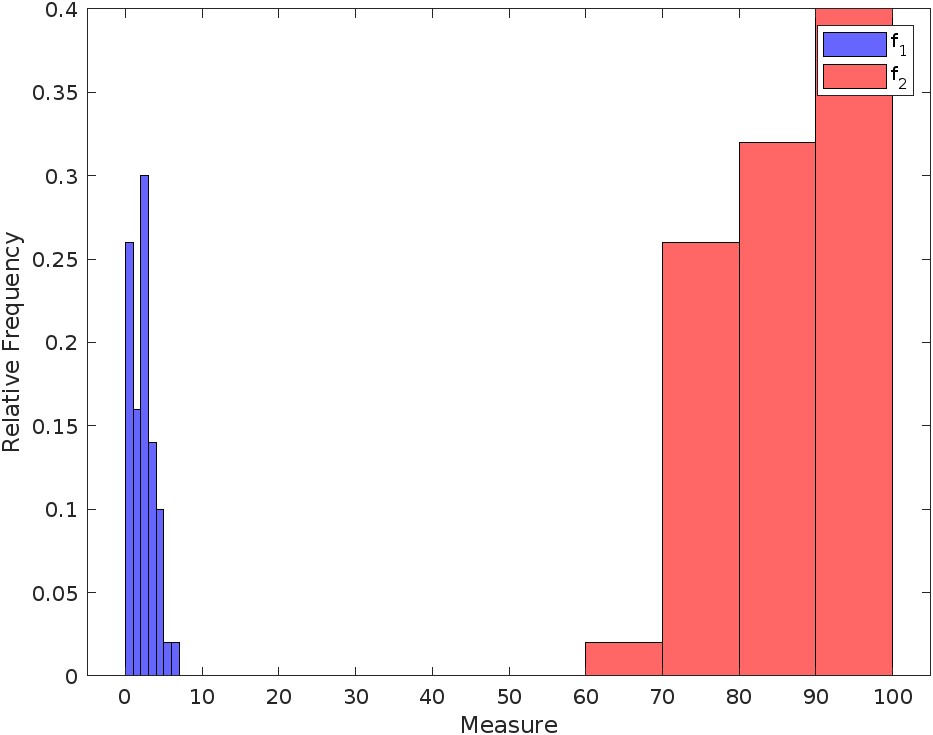}} 
    \caption{(a) Example of dissimilar curves. (b) Examples of similar curves. (c) Training data histogram of $f_1$ and $f_2$ for the dissimilar curves. (d) Training data histogram of $f_1$ and $f_2$ for the similar curves. (e) Test data histogram of $f_1$ and $f_2$ for the dissimilar curves. (f) Test data histogram of $f_1$ and $f_2$ for the similar curves.}
    \label{figure19}
\end{figure}

There are several mathematical models of drug dissolution; see, for example, an important model of drug dissolution is the logistic curve $f=\frac{100}{1+\exp(-c(t-6))}$ for some release rate $c>0$, see \cite{Pourmohamad2023}.  We follow \cite{Pourmohamad2023} by using the logistic curve at the sampling time
points, t = 5, 10, 15, 20, 30, 60, 90, and 120 minutes. To generate a number of similar and dissimilar profiles,  we use release rates $c_1=.01+.001*z$ and $c_2=.03+.001*z$ where $z$ are generated randomly and independently from the standard normal distribution. We can see the simulated curves in Figure \ref{figure19} for (a) dissimilar and (b) similar curves. In addition, we have the histogram for $f_1$ and $f_2$ for the dissimilar (c) and similar (d) curves in the training set.  Figure also shows the histogram for $f_1$ and $f_2$ for the dissimilar (e) and similar (f) curves in the test set.

\begin{figure}[H]
    \centering
    \includegraphics[width=7cm]{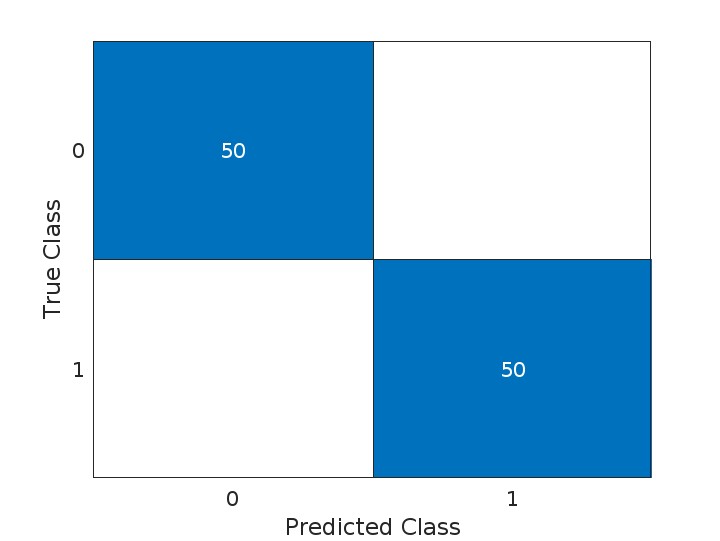}
    \caption{Confusion matrix shows the true predicted positives and the true predicted negatives without resulting in false positives or negatives when comparing 50 similar and 50 dissimilar pairs of curves.}
    \label{fig:figure20}
\end{figure}

To detect the similarity of any two drug dissolution curves, we use a Siamese CNN in which two parallel CNN's final layers are inputs to a cross-entropy measure of the two input images. Some minor changes are made to the overall CNN architecture, in which we avoid using batch normalization and change the average pooling layer to a max pooling layer. In the end, there is a dense layer with $28^2$ hidden nodes. Also, the weights are initialized by sampling from a normal distribution of mean zero and a standard deviation of 0.01. Following \cite{koch2015siamese}, we use the cross-entropy in the output layer to identify the similarity between the images of the dissolution curves. The test resutls can be seen in the confusion matrix in Figure \ref{fig:figure20}.

It is important to note that the number of hidden nodes in the last layer helps in convergence. We notice that it is possible to use one hidden node; however, convergence is not consistently guaranteed. The number of nodes of the hidden layer therefore should be considered an important hyperparameter. Furthermore, below $28^2$ nodes, there was no considerable change in convergence time.

\section{Real-Life Application: Detecting Parkinson's Disease}
Parkinson's disease is a progressive neurodegenerative disorder that results in motor and non-motor symptoms such as tremors, rigidity, and impaired movement control. Detecting Parkinson's disease involves a thorough physical examination to assess motor skills, reflexes, muscle strength, and coordination, and searching for characteristic signs of Parkinson's disease.

\begin{figure}[H]
    \centering
    \subfigure[]{\includegraphics[width=6.25cm]{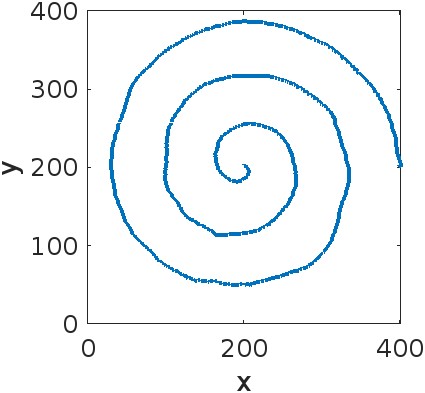}} 
    \subfigure[]{\includegraphics[width=6cm]{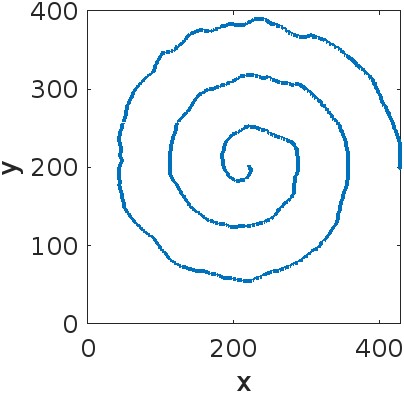}}
    \caption{(a) A spiral drawing of a normal subject. (b) A spiral drawing of a Parkinson's patient. }
    \label{fig:figure21}
\end{figure}

Our technique could be successfully applied to detect Parkinson's disease using motor tests. We use a dataset introduced by \cite{isenkul2014improved,isenkul2014data} in which 62 Parkinson's patients and 15 healthy subjects draw a spiral curve on a tablet. The original test was divided into three parts: a static test, a dynamic test, and a circular motion test. In the static test, subjects draw a certain fixed shape. In the dynamic test, the subjects draw a shape that disappears and appears at certain times, so the subjects need to memorize their location and continue drawing. In the circular motion test, subjects draw a circle at a red reference point on the tablet. Here, we use the data set for the static drawing test of Parkinson's patients by \cite{isenkul2014data}, which contains at each time stamp the positions $X$, $Y$, $Z$ of the drawing, the pressure that we call $P$, and the grip angle that we call $A$, see Figure \ref{fig:figure21}.

\cite{akyol2017study} used three ML/DL techniques, random forest, logistic regression, and an artificial neural network, to classify subjects into Parkinson's and non-Parkinson's. While the original data set had only 77 subjects, \cite{akyol2017study} used tens of thousands of instances. Among the three ML/DL techniques, the artificial neural network showed 100\% precision in all cases. 

\cite{kamble2021digitized} used logistic regression, random forest, support vector machines, and K-Nearest Neighbors using unbalanced data from 25 patients and 15 controls. Logistic regression was the best model using the AUC. With the same idea, \cite{thakur2023automated} used logistic regression, a support vector machine, and a restricted Boltzmann machine followed by a neural network for the same data set. The restricted Boltzmann machine with the neural network model achieved an accuracy of 95\%. See also a review on the use of ML/DL techniques to detect Parkinson's disease \cite{mei2021machine}.

\begin{figure}[H]
    \centering
    \subfigure[]{\includegraphics[width=6cm]{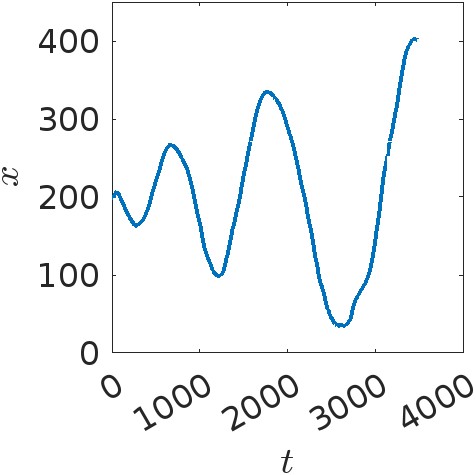}} 
    \subfigure[]{\includegraphics[width=6cm]{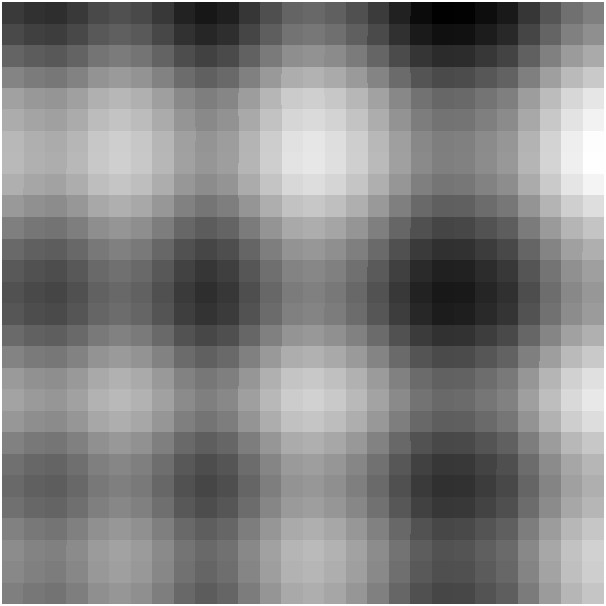}}
    \subfigure[]{\includegraphics[width=6cm]{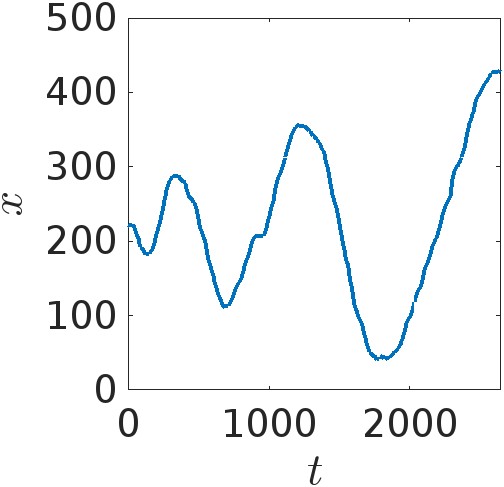}}
    \subfigure[]{\includegraphics[width=6cm]{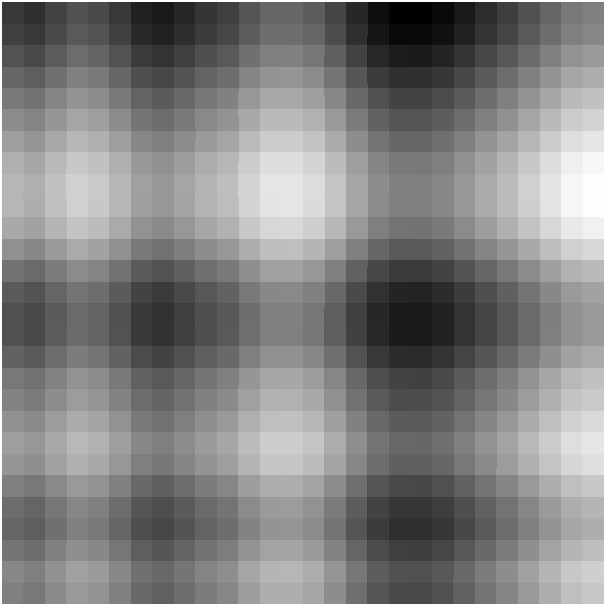}}
    \caption{Curves with their respective image transformation. (a) Example of a control subject's $x$ spiral component time series. (b) The 28 by 28 image corresponds to the curve in (a). (c) Example of a Parkinson's patient's $x$ spiral component time series. (d) The 28 by 28 image corresponds to the curve in (c). }
    \label{fig:figure22}
\end{figure}

Regarding the data set used, due to the small size of the data, we use augmentation (see \cite{james2013introduction} for more details on data augmentation) to make a larger training and testing data set. We made a tensor, in which the third dimension represents the fixed signals of $X$ and $Y$, while we checked a combination of $Z$, the pressure measurement $P$, and the grip angle $A$. So, we have 8 tensors with size 1024 by 1024 by 2 plus the combination of the last 3 variables to determine which has the best results. Using MATLAB data augmentation (see \cite{MATLAB:2022} functions for more details), we made a random translation of $X$ within $[-25,25]$, random reflections, and translation of $Y$ within $[-50,50]$, adding a random translation for $Z$ within $[-.5,.5]$. We also performed a random translation of the pressure within $[0, 50]$ and a random translation of the angle within $[0, 25]$. Data augmentation was performed four times more for the control class of the set since there are only 15 control images compared to the 61 case images. By doing that augmentation, we prepared training data of 300 control images and 305 disease images. In combination, we split it into 80\% for the training and validation set, and the rest 20\% is used in testing. See examples of Parkinson's and control patients in Figure \ref{fig:figure22}.

\begin{figure}[H]
    \centering
    \includegraphics[width=7cm]{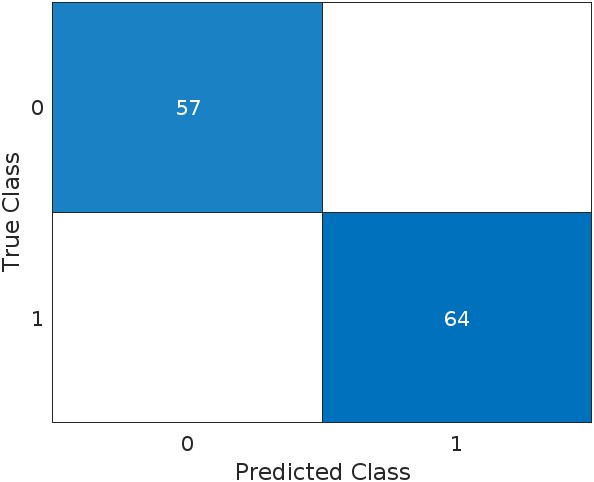}
    \caption{Confusion matrix shows the true predicted positives and the true predicted negatives without resulting in false positives or negatives when using $xy$ combination only.}
    \label{fig:figure23}
\end{figure}

Our method gave 100\% validation and testing accuracy for all of the combinations of the features  $X$, $Y$, $Z$ coordinates, the pressure $P$, and the griping angle of $A$. Using the simplest model with $X$ and $Y$ gives the confusion matrix in Figure \ref{fig:figure23}.

\section{Discussion}
In this paper, we tested our new method of functional data learning using convolutional neural networks (CNN) with various examples and applications. CNN performance was very close to perfect, as evidenced by the test curves of the functional cases in both regression and classification problems. We also see that the variation starts to appear in practical cases such as the chaotic system of Lorentz's attractor and when estimating transmission rates from epidemic curves of the SIR system. Also, we found that training the CNN with noisy data improves CNN performance. These results show that in both cases the new method is robust to noise and would handle different cases of functional data. While some of the p-values of the results show slopes and intercepts that are statistically significant from one to zero, respectively, their estimates and the correlation coefficients show close values to one and zero, indicating an excellent effect size. This conflict might be the result of using large data sizes when testing the trained CNN.

On the practical side, a pre-trained CNN can be used in some of the applications. A pre-trained CNN could estimate Lyapunov exponents and assess the stability of some systems. For example, it could be used in the medical field to determine the stability of human motion or walking gait. This methodology provides a more practical approach in which, with moderate noise, the CNN performs well and is approximately 2 orders of magnitude faster. As such, the measured data can be used as input without the need to filter before using the CNN since it is robust to noise. A CNN pre-trained on epidemic curves can be used to estimate the transmission rate or directly estimate the basic reproduction number $R_0$ and discern exponential growth from algebraic growth. 

When it comes to classification problems, the new method always gave an accuracy of 100\%. We tested the new method for the classification of curves according to their monotonicity and curvature, as well as their type of growth. Furthermore, we simulated drug dissolution profiles to train a Siamese CNN which accurately managed to determine their similarity and dissimilarity. Finally, using real-life data, CNN trained with functional motion data of a few cases of Parkinson's disease and even fewer controls discerned cases with an accuracy of 100\%.   

\section{Conclusion}
In this paper, we show a simple method to convert any curve into an image. Using convolution neural networks (CNN), we trained on a number of those images, together with a validation set of images in various regression and classification problems. The same technique could be used for regression and classification problems in gene analysis and other medical sciences. Other areas where one can explore are multiple output problems in which several parameters are estimated, or classification and regression problems are combined to find the type of curve and estimate its parameters. Extension of the method to allow other types of kernel to produce those images might be a viable extension to the main idea in this paper. However, the presented technique might require large functional data that we could not find at the time of writing this paper. In that case, we had to perform a data augmentation. New methods for efficient functional synthetic data might be needed to handle small functional data learning problems.  

\section*{Acknowledgment}
This work was supported by the U.S. Department of Defense Manufacturing Engineering Education Program (MEEP) under Award No. N00014–19-1–2728. 

The authors also thank Salman Rahman and Harrinson Arrubla for their early discussions and trials on this project. The authors did not receive any funding for the research performed in this article. The authors declare that they have no conflict of interest.

\section*{Data Availability Statement}
This study did not involve the creation, collection, or generation of new data. The study mainly relied on publicly available data previously published.

\section*{Conflict of interest}
The authors declare that they have no conflict of interest.

\section*{Code details}
The codes used in this study can be accessed at \url{https://github.com/jesusgl86/FDAP01}.

\section*{References}
\bibliographystyle{jphysicsB}

\end{document}